\documentclass[lettersize,journal]{IEEEtran}
\usepackage{amsmath,amsfonts}
\usepackage{algorithmic}
\usepackage{array}
\usepackage[caption=false,font=normalsize,labelfont=sf,textfont=sf]{subfig}
\usepackage{textcomp}
\usepackage{stfloats}
\usepackage{url}
\usepackage{verbatim}
\usepackage{graphicx}
\usepackage{amssymb}
\usepackage{arydshln}
\usepackage{float}
\usepackage{multirow}
\usepackage{footnote}
\usepackage{color} 
\usepackage[table]{xcolor}
\usepackage{colortbl}
\usepackage[colorlinks,citecolor=blue,linkcolor=blue]{hyperref}
\usepackage{lineno}
\usepackage{threeparttable}
\usepackage{fancyhdr}

\makeatletter
\newcommand{\coloredcite}[1]{{\color{blue}\cite{#1}}}
\makeatother

\def\BibTeX{{\rm B\kern-.05em{\sc i\kern-.025em b}\kern-.08em
    T\kern-.1667em\lower.7ex\hbox{E}\kern-.125emX}}
\usepackage{balance}

\begin{document}
\title{A Robust and Efficient Boundary Point Detection Method by Measuring Local Direction Dispersion}
\author{Dehua Peng, Zhipeng Gui,\textit{ Member, IEEE}, Jie Gui,\textit{ Senior Member, IEEE}, Huayi Wu
\thanks{This work was supported by National Natural Science Foundation of China (No. 42090010, No. 41971349, No. 41930107) and the Fundamental Research Funds for the Central Universities, China (No. 2042022dx0001). \textit{(Corresponding author: Zhipeng Gui.)}

Dehua Peng is with the School of Remote Sensing and Information Engineering, Wuhan University, Wuhan 430079, China, and also with the State Key Laboratory of Information Engineering in Surveying, Mapping and Remote Sensing, Wuhan University, Wuhan 430079, China. E-mail: pengdh@ whu.edu.cn. 

Zhipeng Gui is with the School of Remote Sensing and Information Engineering, Wuhan University, Wuhan 430079, China, and also with the Collaborative Innovation Center of Geospatial Technology, Wuhan University, Wuhan 430079, China. E-mail: zhipeng.gui@whu.edu.cn.

Jie Gui is with the School of Cyber Science and Engineering, Southeast University, Nanjing 211189, China, also with the Engineering Research Center of Blockchain Application, Supervision and Management, Southeast University, Nanjing 211189, China, and also with Purple Mountain Laboratories, Nanjing 210000, China. E-mail: guijie@seu.edu.cn.
 
Huayi Wu is with the State Key Laboratory of Information Engineering in Surveying, Mapping and Remote Sensing, Wuhan University, Wuhan 430079, China, and also with the Collaborative Innovation Center of Geospatial Technology, Wuhan University, Wuhan 430079, China. E-mail: wuhuayi@ whu.edu.cn.

}}

\maketitle

\thispagestyle{fancy}
\lhead{}
\lfoot{}
\cfoot{\small{Copyright © 2025 IEEE. Personal use of this material is permitted. However, permission to use this material for any other purposes must be obtained from the IEEE by sending an email to pubs-permissions@ieee.org.}}
\rfoot{}

\begin{abstract}
Boundary point detection aims to outline the external contour structure of clusters and enhance the inter-cluster discrimination, thus bolstering the performance of the downstream classification and clustering tasks. However, existing boundary point detectors are sensitive to density heterogeneity or cannot identify boundary points in concave structures and high-dimensional manifolds. In this work, we propose a robust and efficient boundary point detection method based on Local Direction Dispersion (LoDD). The core of boundary point detection lies in measuring the difference between boundary points and internal points. It is a common observation that an internal point is surrounded by its neighbors in all directions, while the neighbors of a boundary point tend to be distributed only in a certain directional range. By considering this observation, we adopt density-independent K-Nearest Neighbors (KNN) method to determine neighboring points and design a centrality metric LoDD using the eigenvalues of the covariance matrix to depict the distribution uniformity of KNN. We also develop a grid-structure assumption of data distribution to determine the parameters adaptively. The effectiveness of LoDD is demonstrated on synthetic datasets, real-world benchmarks, and application of training set split for deep learning model and hole detection on point cloud data. The datasets and toolkit are available at: \url{https://github.com/ZPGuiGroupWhu/lodd}.
\end{abstract}


\section{Introduction}\label{sec1}
\begin{figure}[t]
\centering
\includegraphics[width=1\linewidth]{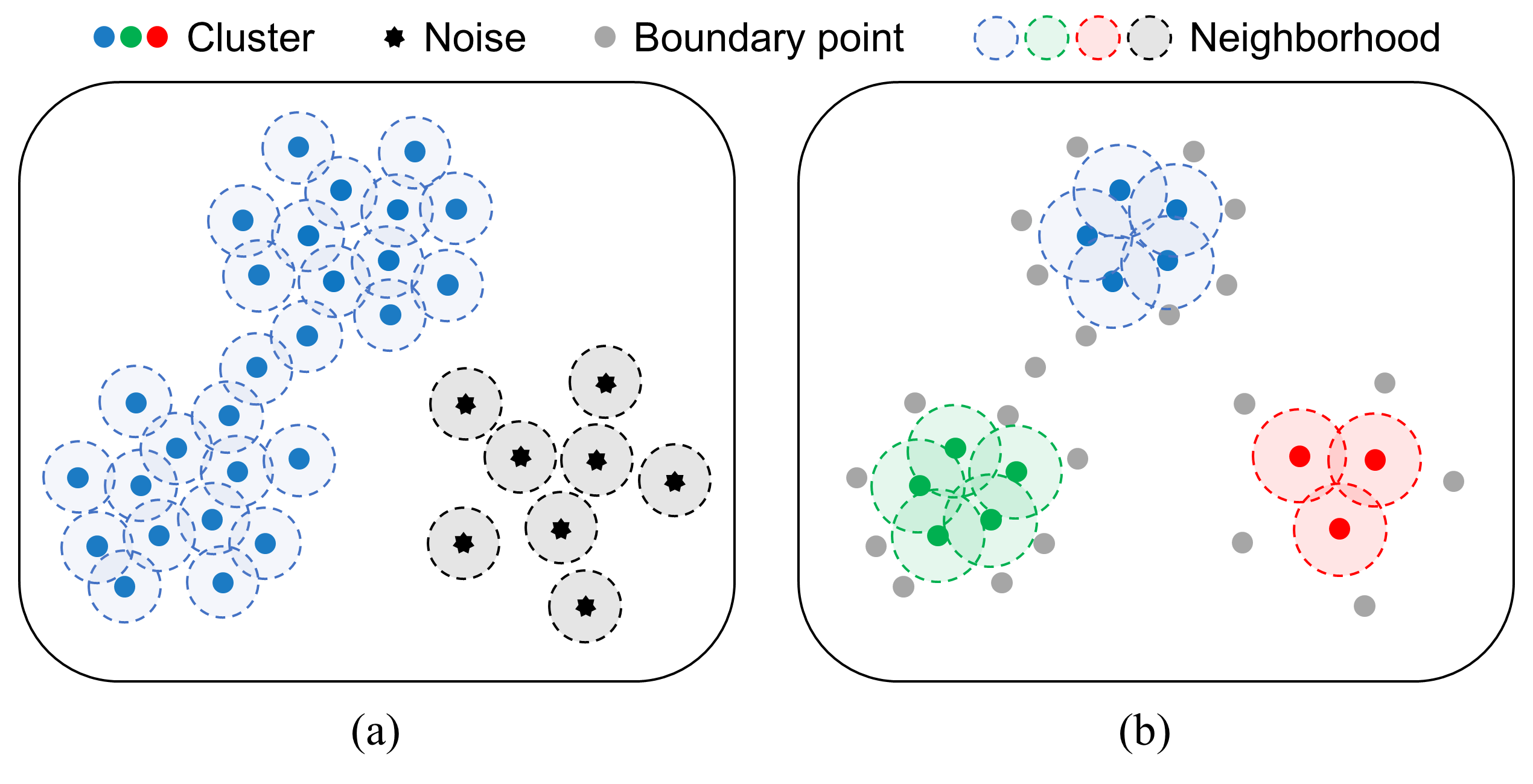}
\caption{An example on a toy dataset that reveals the performance enhancement of the DBSCAN algorithm by boundary point extraction. (a) Weak connectivity and density heterogeneity impede the effectiveness of DBSCAN. (b) DBSCAN can separate the weakly connected clusters and identify the sparse cluster by peeling out the boundary points.}
\label{fig1}
\end{figure}
\IEEEPARstart{B}{OUNDARY} point detection has been widely used in object edge extraction, point cloud processing, and common machine learning tasks such as clustering and classification \coloredcite{r1, r2, r3}. The boundary point has a critical impact on the effectiveness of classification and clustering methods. The essence of classification and clustering lies in identifying the aggregation patterns hidden behind the data \coloredcite{r4, r5}. An aggregation pattern typically consists of internal points and boundary points. The internal points are densely located in the core of a group, while the boundary points lie in the external contour and define the range and extrema \coloredcite{r6}. When data points have similar features, the clusters would overlap each other and the boundaries would be fuzzy. In this case, fuzzy boundary points introduce considerable difficulty for machine learning tasks and serve as the primary source of classification and clustering errors \coloredcite{r7}. As a typical classifier, a Support Vector Machine (SVM) seeks an optimal decision hyperplane to segregate the data samples by solving a dual problem with Lagrange multipliers\coloredcite{r9}. However, when the clusters are nonlinearly distributed with more complex boundaries, hard-margin SVM must be extended to soft-margin SVM by introducing more constraints based on the primal optimal problem\coloredcite{r10}. Similar situations can occur with Maximum Margin Clustering (MMC), which utilizes the maximum margin principle popularized by supervised SVM. Although it has achieved success in video motion recognition \coloredcite{r12, r8, r13, r54}, fuzzy boundaries still lead to significant errors. In this case, detecting and removing the boundary points can make the clusters linearly separable and enables SVM to find a good separator more easily. Connectivity-based clustering is another important branch of cluster analysis that excels in identifying arbitrary shaped clusters. Density-based Spatial Clustering of Applications with Noise (DBSCAN) leverages circular neighborhoods to connect data points whose densities exceed a threshold \coloredcite{r11}. However, its effectiveness is affected by weak connectivity and density heterogeneity. Fig. \ref{fig1}\textcolor{blue}{(a)} presents a toy dataset including two weakly connected clusters and a low-density cluster. DBSCAN connects these two weakly connected clusters as a whole and misidentifies the sparse cluster as noise. Peeling out the boundary points can enlarge the inter-cluster gaps and make the reasonable range for setting the neighborhood radius larger (Fig. \ref{fig1}\textcolor{blue}{(b)}). 

The impact of boundary points has received extensive attention, and a number of boundary point detectors have been developed. Current boundary point detectors are mainly categorized into density-based methods and geometry-based methods. Density-based methods take into account that boundary points often have lower local density compared to internal points \coloredcite{r14}. They have been widely used due to the straightforward idea and strong adaptability. However, density-based methods are susceptible to the influence of density heterogeneity and may easily misidentify sparse clusters as boundary points. Geometry-based methods can better cope with density heterogeneity by considering geometric characteristics, such as curvature \coloredcite{r18} and gravitational direction \coloredcite{r20}. Nonetheless, the computation of geometric characteristics is often computationally intensive and may not perform well in identifying boundary points for concave structured clusters.

In this work, we propose a novel boundary point detector LoDD by measuring local direction dispersion. To validate the effectiveness, we compared LoDD with five mainstream baselines on a total of 19 datasets derived from different fields, including three synthetic datasets, 14 real-world benchmarks, and two 3-D point clouds. LoDD demonstrates notable advantages against the competitors. Specifically, it improved the ACC (ACCuracy) score of K-means by an average of 8.71\% and the NMI (Normalized Mutual Information) score by 6.16\% on 14 real-world benchmarks. In terms of time efficiency, LoDD runs 281, 36, 10, and 3 times faster than DCM, ROBP, NC, and LDIV implemented in Python on 100k samples, respectively. It can even achieve a greater computing advantage in the MATLAB environment, being 610 times more efficient than DCM. The major contributions of our work can be summarized as follows:
\begin{itemize}
\item A centrality metric is developed to detect boundary points via the eigenvalues of local covariance matrix. It can cope with density heterogeneity and identify boundary points of concave structures and high-dimensional manifolds. Besides, the time efficiency is improved using the matrix trace to bypass expensive computation of eigenvalues.
\item An adaptive parameter setting strategy is designed based on a grid-structure assumption. It does not require extra computational overhead by estimating the upper and lower bounds of the number of boundary points.
\item The potential applications in machine learning are explored. LoDD enhanced unsupervised clustering accuracy, and improved the performance of deep learning model by guiding training set split in an image classification task. It also achieved the highest accuracy for detecting boundaries and holes in two point cloud data. 
\end{itemize}

The remainder of this paper is organized as follows. Section \ref{sec2} reviews the related literature. Section \ref{sec3} introduces the preliminaries and motivation of our method, and Section \ref{sec4} presents the detailed algorithm procedure. Section \ref{sec5} develops an adaptive method for parameter settings. In Section \ref{sec6}, we demonstrate the effectiveness on challenging synthetic and real-world datasets and apply it to point cloud data. Finally, our conclusions are drawn in Section \ref{sec7}.
\section{Related Work}\label{sec2}
Existing boundary detectors can be summarized into two major categories, namely density-based and geometry-based methods. Internal points are located in the high-density core regions, while boundary points sit near the margins or free pattern space of the densely distributed classes with a relatively lower density. Considering the density difference, a BOundaRy point DEtectoR (BORDER) was proposed. It depicts the point density using Reverse K-Nearest Neighbors (RKNN), which identifies the points with a lower RKNN as boundary points \coloredcite{r14}. However, it only counts the number of RKNN, but ignores the influence brought by distance between the query point and its neighbors. Border-Peeling Clustering (BPC) designs a density influence factor to detect and peel the boundary points iteratively. It employs Euclidean distance and the Gaussian kernel to estimate the combined density influence of RKNN \coloredcite{r15}. Instead of an inherent neighborhood parameter, self-adaptive neighborhood border peeling clustering uses natural nearest neighbors to adaptively obtain the neighborhood parameter, which can generate a different neighborhood for each point according to different shapes of the data \coloredcite{r16}. To improve the density influence factor, RObust Border-Peeling (ROBP) uses the Cauchy kernel and a linkage criterion based on the shared neighborhood information \coloredcite{r17}, while Local Density Influence Values (LDIV) computes the local relative density by taking RKNN as weights \coloredcite{r49}. However, density-based methods suffer from the issue of density heterogeneity. 

Geometry-based methods hence provide another perspective. Based on the intuitive observation that boundary points sit on one side of the cluster surface, Border-Edge Pattern Selection (BEPS) constructs the approximated tangent hyperplane of a class surface and measure the directional consistency between the normal vector and vectors formed by the query point and its KNN \coloredcite{r18, r19}. Local Gravitation Clustering (LGC) adopts centrality and coordination to measure the directional consistency between the local attractive forces and mean-shift directions of neighbors \coloredcite{r20}. Nonetheless, BEPS and LGC might be invalid for clusters of non-convex structure. To handle non-convex structure, Level Set Boundary Description (LSBD) first fits the data distribution using kernel density estimation without any assumptions, and then determines the enclosed boundary surface \coloredcite{r21}. This prototype-based method is supervised and requires a time-consuming training process. Besides, an efficient boundary detection algorithm namely Negative Componnet (NC) was presented \coloredcite{r22}, in which the neighborhood relationships of each point are encoded as a linear data representation in an affine space. The reverse unreachability defined by the number of negative components of the representation matrix is exploited to identify the boundary and outlier points \coloredcite{r23}. But when the local covariance matrix is not invertible, the representation matrix is unable to be solved. Recently, we proposed a novel boundary-seeking clustering algorithm by calculating the local Direction Centrality Metric (DCM) \coloredcite{r24}. Its core idea is to measure the uniformity of the directional distribution of KNN. Specifically, DCM is defined as the variance of angles formed by the query point and its KNN in 2-D space, and the variance of simplex volume in higher dimensions. Whereas, the calculation of DCM is computationally intractable since the number of simplexes increases exponentially with dimensions. Therefore, a density-independent and efficient boundary point detection method for irregular shaped structures and complex manifolds is highly desired.
\begin{figure}[t]
\centering
\includegraphics[width=0.9\linewidth]{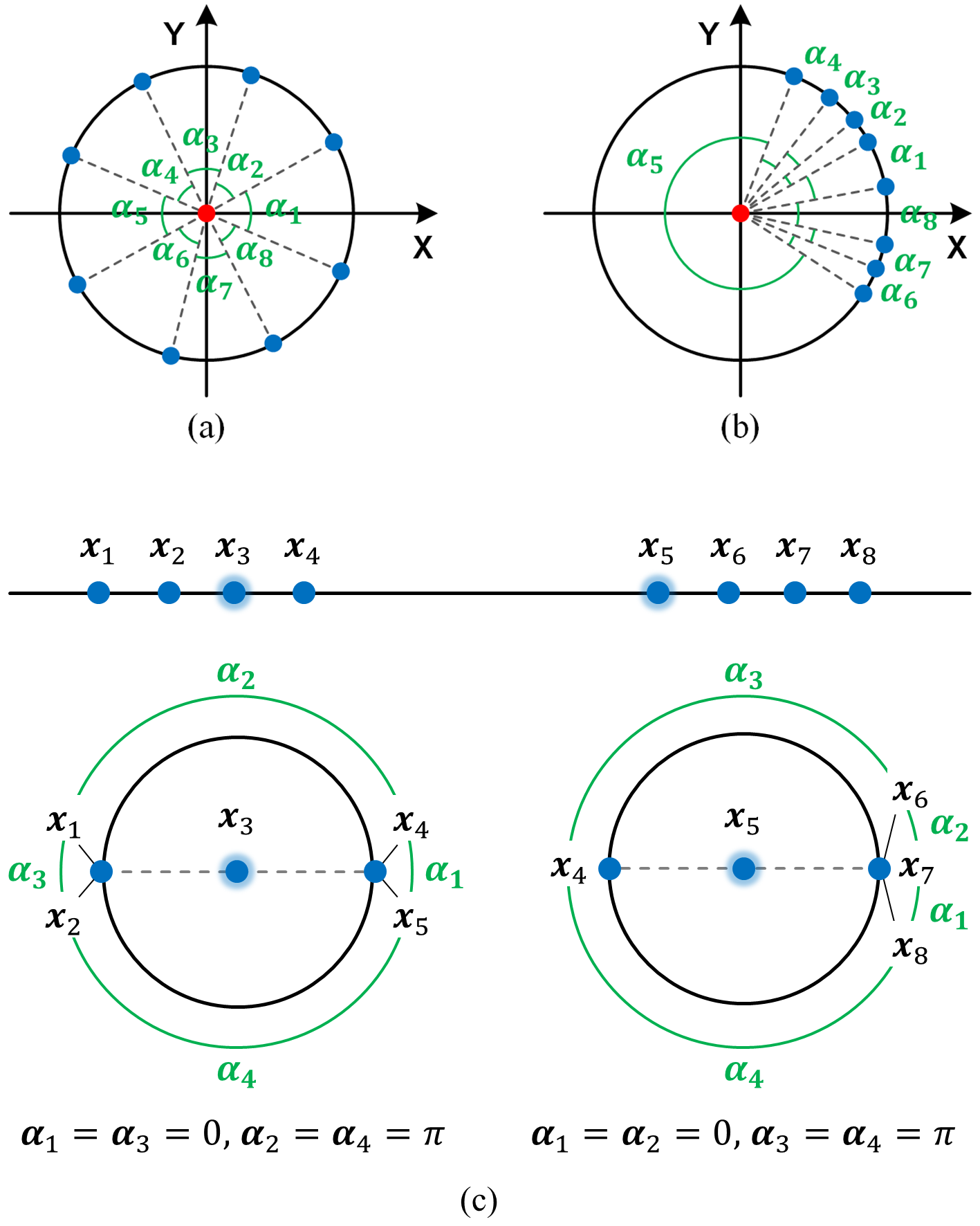}
\caption{Ability and limit of Direction Centrality Metric (DCM). (a) The central angles are approximately equal when KNN are distributed uniformly on the unit circle. (b) The central angles differ a lot when KNN lie in a smaller range of direction. (c) A 2-D toy data illustrates that DCM fails to reflect the uniformity of KNN distribution when handling manifold clusters.}
\label{fig2}
\end{figure}

\section{Preliminaries and Motivation}\label{sec3}
Previous researches \coloredcite{r24,r43,r44} demonstrated that Direction Centrality Metric (DCM) has a robust performance for handling low-dimensional data of complex structures. However, DCM is limited by excessive computational cost in high-dimensional space. This paper intends to follow such a simple and intuitive idea, and evades the computable problem through an alternative calculation strategy. In this section, we first provide a brief review of DCM. Then, we propose a novel centrality metric to measure the local direction dispersion.
\subsection{Direction Centrality Metric}
As described in Section \ref{sec2}, the distribution uniformity of KNN can reflect the centrality of a query point. An intuitive example of 2-D points is given in Fig. \ref{fig2}. Specifically, we search for the eight nearest neighbors of the current query point ($k=8$) and project them onto the unit circle centered at the query point. We do not take the distances of KNN into account. As shown in Fig. \ref{fig2}\textcolor{blue}{(a)}, the eight central angles are approximately equal when the KNN are evenly distributed. While when the KNN are located in a smaller directional range as illustrated in Fig. \ref{fig2}\textcolor{blue}{(b)}, angle $\alpha_5$ is much larger than the other seven angles. Based on this observation, we explicitly define DCM as the variance of angles in 2-D space

\begin{equation}
DCM = \frac{1}{k}{\sum_{i = 1}^{k}( \alpha_{i} - \frac{2\pi}{k})^{2}}
\label{eq1}
\end{equation}
where $\alpha_1, \alpha_2,..., \alpha_k$ denote the $k$ central angles formed by the query point and its KNN, and the condition ${\sum_{i=1}^{k}\alpha_{i}}=2\pi$ holds in 2-D space. DCM reaches the minimum 0 if and only if all the angles are equal, which means that the KNN are evenly distributed in all directions. DCM can be maximized as $4(k-1)\pi^{2}/{k^{2}}$ when one of these angles equlas to $2\pi$ and the remaining are 0. Such an extreme situation happens when all KNN are distributed in the same direction.

DCM is robust to the cluster shape and density heterogeneity. However, the concept of DCM must be extended to the variance of simplex volumes in higher-dimensional spaces due to the computable problem of central angles. Subsequently, the number of simplices grows exponentially as dimensions, leading to low computational efficiency. Besides, DCM cannot accurately reflect the uniformity of the KNN distribution when dealing with manifold clusters. We give an example on a 2-D toy data in Fig. \ref{fig2}\textcolor{blue}{(c)} to explain this point. It contains eight collinear points, where $\mathbf{x}_1$ to $\mathbf{x}_4$ form a manifold cluster and $\mathbf{x}_5$ to $\mathbf{x}_8$ form another manifold cluster. By setting $k=4$, $\mathbf{x}_1$, $\mathbf{x}_2$, $\mathbf{x}_4$, and $\mathbf{x}_5$ are the 4-nearest neighbors of $\mathbf{x}_3$, while $\mathbf{x}_4$, $\mathbf{x}_6$, $\mathbf{x}_7$, and $\mathbf{x}_8$ are that of $\mathbf{x}_5$. To calculate DCM, their KNN are mapped onto their respective unit circles and form four central angles. The angles formed by the KNN of $\mathbf{x}_3$ are 0, 0, $\pi$, and $\pi$, which are the same with $\mathbf{x}_5$. It means that $\mathbf{x}_3$ and $\mathbf{x}_5$ have the same DCM value and centrality. However, this is unreasonable because there are two points on both the left and right sides of $\mathbf{x}_3$, while $\mathbf{x}_5$ has one point on the left and three points on the right. The imbalanced distribution of KNN of $\mathbf{x}_5$ is not reflected by the angle variance. In this work, we intend to follow the idea of DCM, but refine its definition using projection variance to handle manifolds of any dimension in an efficient manner.

\begin{figure}[t]
\centering
\includegraphics[width=0.97\linewidth]{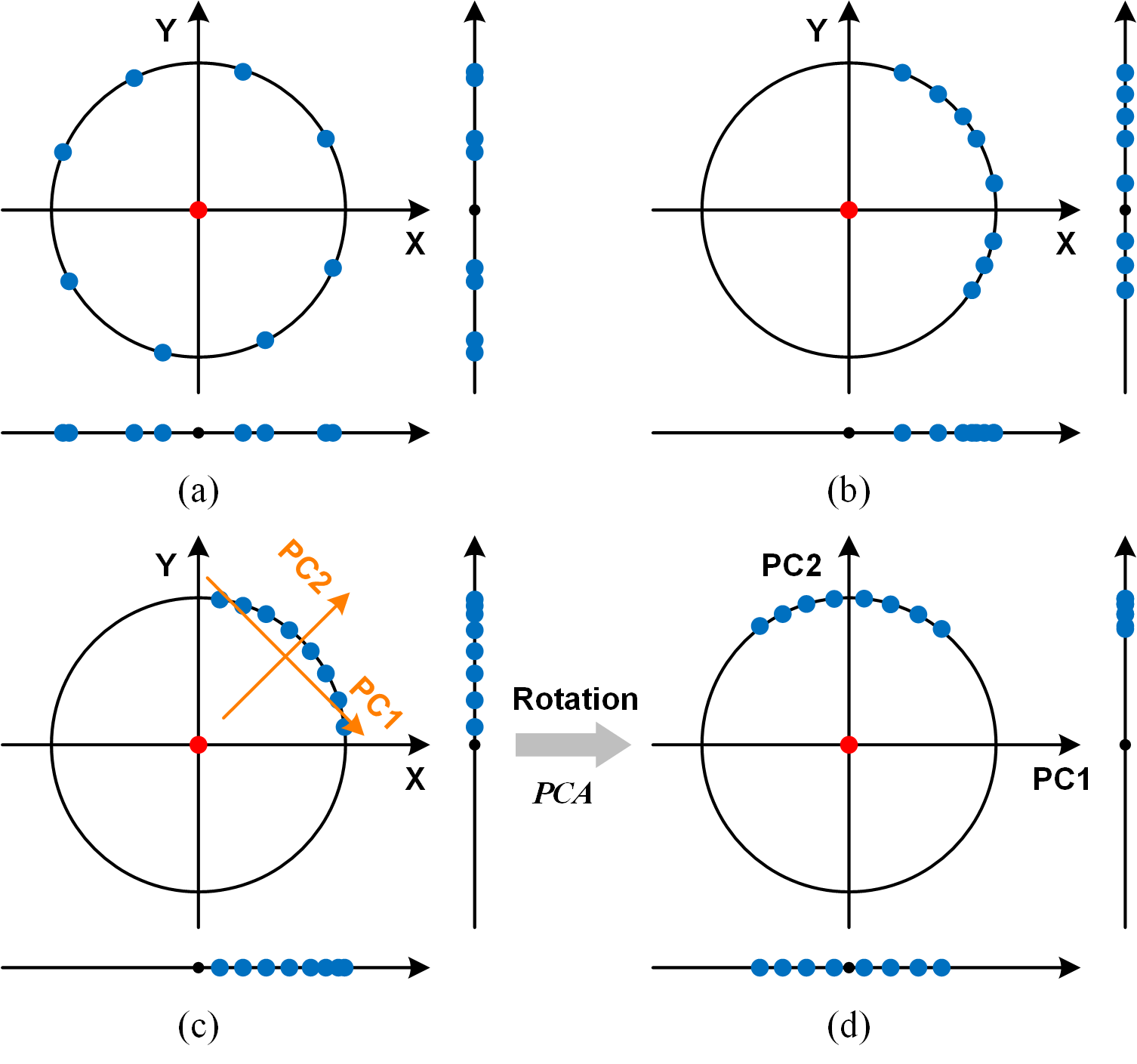}
\caption{Illustration of how the projection variances of KNN indicate the centrality. (a) The case when KNN are distributed uniformly on the unit circle, in which the projections are located dispersedly on the axes. (b) The case when KNN are distributed on a smaller range of the unit circle, in which the projections are located compactly on the X axis. (c) PCA rotates the KNN to the varimax direction. (d) The distribution projections on the axes of KNN after PCA rotation.}
\label{fig3}
\end{figure}

\subsection{Projection Variance and Centrality}
Projection variance indicates the distribution uniformity of KNN. Taking 2-D points as an example in Fig. \ref{fig3}, the projections are dispersed on both axes when KNN are uniformly located on the unit circle (Fig. \ref{fig3}\textcolor{blue}{(a)}). Whereas, when KNN are situated in a smaller range, the projections are relatively concentrated on the X axis (Fig. \ref{fig3}\textcolor{blue}{(b)}). 

Nonetheless, the original projection variances cannot represent the real distribution of KNN. Let us assume that the KNN points are collectively rotated by an angle along the unit circle, their relative positions remain unchanged, but the projection variances vary. Hence, we first use Principal Component Analysis (PCA) to rotate the KNN in the varimax direction as shown in Fig. \ref{fig3}\textcolor{blue}{(c)}, and then measure the projection variances on the rotated axes (Fig. \ref{fig3}\textcolor{blue}{(d)}). That is because PCA can maximize the difference between the maximum and minimum projection variances. The PCs are represented by orthogonal lines in the high-dimensional variable space, intersecting at the center of mass of KNN due to the mean-centering process. In PCA, the eigenvalues of the covariance matrix are equal to the projection variances on the PCs. In such a scenario, if the variances are larger and more similar, the probability of the query point being a boundary point is lower.

Below, we illustrate the relationship between KNN projection variances and centrality of query point through theoretical analysis in 2-D space. Given the coordinates of a query point $(a_0, b_0)$ and its KNN $\{(a_1, b_1),(a_2, b_2),\ldots,(a_k, b_k)\}$, we map the KNN onto the unit circle with the center of $(a_0, b_0)$, and the coordinates on the unit circle can be calculated as

\begin{equation}
\left\{ {\begin{matrix}
{\hat{a_{i}} = \frac{a_{i} - a_{0}}{\sqrt{\left( a_{i} - a_{0} \right)^{2}~ + ~\left( b_{i} - b_{0} \right)^{2}}}} \\
~\\
{\hat{b_{i}} = \frac{b_{i} - b_{0}}{\sqrt{\left( a_{i} - a_{0} \right)^{2}~ + ~\left( b_{i} - b_{0} \right)^{2}}}}
\end{matrix}} \right.
\label{eq2}
\end{equation}
where $i$ ranges from 0 to $k$.

We calculate the population covariance matrix $\mathbf{C} \in \mathbb{R}^{2 \times 2}$ of $\{({\hat{a_{1}},\hat{b_{1}}}),( {\hat{a_{2}},\hat{b_{2}}}),\ldots,( \hat{a_{k}},\hat{b_{k}}) \}$, and solve its eigenvalues as $\lambda_1$ and $\lambda_2$. Suppose the coordinates of KNN after PCA rotation as $\{(x_1, y_1),(x_2, y_2),\ldots,(x_k, y_k)\}$, then we have

\begin{equation}
\left\{ {\begin{matrix}
{\lambda_{1} = \frac{1}{k}{\sum_{i = 1}^{k}\left( x_{i} - \overline{x} \right)^{2}}} \\
~\\
{\lambda_{2} = \frac{1}{k}{\sum_{i = 1}^{k}\left( y_{i} - \overline{y} \right)^{2}}}
\end{matrix}} \right.
\label{eq3}
\end{equation}
where $\overline{x} = \frac{1}{k}{\sum_{i = 1}^{k}x_{i}}$ and $\overline{y} = \frac{1}{k}{\sum_{i = 1}^{k}y_{i}}$. 

\textbf{Theorem 1:} When the centroid of KNN overlaps with the query point, $\lambda_1+\lambda_2$ reaches its maximum value of 1; while when all KNN perfectly coincide with each other, both $\lambda_1$ and $\lambda_2$ are equal to 0. 

\textbf{Proof:} By adding up $\lambda_1$ and $\lambda_2$, we have
\begin{equation}
\begin{aligned}
&~\lambda_{1} + \lambda_{2} = \frac{1}{k}{\sum\limits_{i = 1}^{k}\left( x_{i} - \overline{x} \right)^{2}} + \frac{1}{k}{\sum\limits_{i = 1}^{k}\left( y_{i} - \overline{y} \right)^{2}}\\
&= \frac{1}{k}{\sum{x_{i}}^{2}} - \frac{2\overline{x}}{k}{\sum x_{i}} + {\overline{x}}^{2} + \frac{1}{k}{\sum{y_{i}}^{2}} - \frac{2\overline{y}}{k}{\sum y_{i}} + {\overline{y}}^{2}\\
&= \frac{1}{k}{\sum{x_{i}}^{2}} - {\overline{x}}^{2} + \frac{1}{k}{\sum{y_{i}}^{2}} - {\overline{y}}^{2}
\end{aligned}
\label{eq4}
\end{equation}

Since KNN lie on the unit circle, the condition ${x_{i}}^{2} + {y_{i}}^{2} = 1$ holds. Then, we can obtain
\begin{equation}
\lambda_{1} + \lambda_{2} = 1 - {\overline{x}}^{2} - {\overline{y}}^{2}
\label{eq5}
\end{equation}

The centroid of KNN is on or inside the unit circle, that means $0 \leq {\overline{x}}^{2} + {\overline{y}}^{2} \leq 1$, so we have
\begin{equation}
0 \leq \lambda_{1} + \lambda_{2} \leq 1
\label{eq6}
\end{equation}

To some extent, $\lambda_{1} + \lambda_{2}$ can indicate the symmetry of KNN. Let us assume that the gravitational force is acting from the query point towards the KNN. If the centroid of KNN is located at the query point, the resultant force of gravitation will be zero, and ${\overline{x}}^{2} + {\overline{y}}^{2}=0$ holds, that means $\lambda_{1} + \lambda_{2}=1$. In the other case, if all KNN overlap on the unit circle, the gravitational forces will point in the same direction, ${\overline{x}}^{2} + {\overline{y}}^{2}=1$ holds, and $\lambda_{1} + \lambda_{2}=0$. $\hfill\square$

\textbf{Theorem 2:} If KNN are uniformly distributed on the unit circle such that the circle is divided into $k$ evenly sized arcs and $k>2$, then $\lambda_{1}=\lambda_{2}=1/2$.

\textbf{Proof:} KNN divide the circle into $k$ equal arcs, so the $k$ central angles formed by the query point and the adjacent KNN are equal to $2\pi/k$. In this case, suppose the polar coordinates of KNN as $\left\{ \left( {1,\theta + 2\pi/k} \right),\left( {1,\theta + 4\pi/k} \right),\ldots,(1,\theta + 2\pi) \right\}$, then we have
\begin{equation}
\left\{ \begin{matrix}
\begin{matrix}
{x_{i} = \cos{\left( {\theta + \frac{2i\pi}{k}}\right)}} \\
~ \\
{y_{i} = \sin{\left( {\theta + \frac{2i\pi}{k}}\right)}}
\end{matrix}
\end{matrix} \right.
\label{eq7}
\end{equation}
where $i$ ranges from 1 to $k$. 

Then, the X coordinate of the centroid can be calculated

\begin{equation}
\begin{aligned}
    &~\overline{x}
    = \frac{\cos\theta}{k}{\sum\limits_{i = 1}^{k}{\cos\left( \frac{2i\pi}{k} \right)}} - \frac{\sin\theta}{k}{\sum\limits_{i = 1}^{k}{\sin\left( \frac{2i\pi}{k}\right)}}\\
    &= \frac{\cos\theta}{2k\sin\left( \frac{\pi}{k}\right)}{\sum\limits_{i = 1}^{k}{2\sin\left( \frac{\pi}{k}\right )\cos\left( \frac{2i\pi}{k}\right)}} \\
    &~~~- \frac{\sin\theta}{2k\sin\left( \frac{\pi}{k}\right)}{\sum\limits_{i = 1}^{k}{2\sin\left( \frac{\pi}{k}\right)\sin\left( \frac{2i\pi}{k}\right)}}\\
    &= \frac{\cos\theta}{2k\sin\left(\frac{\pi}{k} \right)}{\sum\limits_{i = 1}^{k}\left( {\sin\left( \frac{(2i + 1)\pi}{k}\right) - \sin\left( \frac{(2i - 1)\pi}{k}\right)} \right)}\\
    &~~~-\frac{\sin\theta}{2k\sin\left( \frac{\pi}{k} \right)}{\sum\limits_{i = 1}^{k}\left( {\cos\left( \frac{(2i - 1)\pi}{k}\right) - \cos\left( \frac{(2i + 1)\pi}{k}\right)} \right)}\\
    &= \frac{\cos\theta}{2k\sin(\frac{\pi}{k})}\left( {\sin\left( \frac{(2k + 1)\pi}{k}\right) - \sin\left( \frac{\pi}{k}\right)} \right)\\
    &~~~-\frac{\sin\theta}{2k\sin\left( \frac{\pi}{k}\right)}\left( {\cos\left( \frac{\pi}{k}\right) - \cos\left( \frac{(2k + 1)\pi}{k}\right)} \right) = 0
\end{aligned}
\label{eq8}
\end{equation}

Similarly, we can get $\overline{y} = 0$, which means the centroid of KNN lies on the query point when they form $k$ equal angles. We can further obtain

\begin{equation}
\begin{aligned}
&~\lambda_{1} = \frac{1}{k}{\sum\limits_{i = 1}^{k}\left( x_{i} - \overline{x} \right)^{2}} = \frac{1}{k}{\sum\limits_{i = 1}^{k}{x_{i}}^{2}} \\
& = \frac{1}{k}{\sum\limits_{i = 1}^{k}{{\cos}^{2}\left( {\theta + \frac{2i\pi}{k}}\right)}} = \frac{1}{2k}{\sum\limits_{i = 1}^{k}\left( {\cos\left( {2\theta + \frac{4i\pi}{k}}\right)} \right)} + \frac{1}{2}\\
&= \frac{\cos2\theta}{2k}{\sum\limits_{i = 1}^{k}{\cos\left( \frac{4i\pi}{k} \right)}} - \frac{\sin2\theta}{2k}{\sum\limits_{i = 1}^{k}{\sin\left( \frac{4i\pi}{k} \right)}} + \frac{1}{2}\\
&= \frac{\cos2\theta}{4k\sin( \frac{2\pi}{k} )}{\sum\limits_{i = 1}^{k}\left( {\sin\left( \frac{(4i + 2)\pi}{k}\right) - \sin\left( \frac{(4i - 2)\pi}{k} \right)} \right)}-\\
&\frac{\sin2\theta}{4k\sin\left( \frac{2\pi}{k} \right)}{\sum\limits_{i = 1}^{k}\left( {\cos\left( \frac{(4i - 2)\pi}{k}\right) - \cos\left( \frac{(4i + 2)\pi}{k}\right)} \right)} + \frac{1}{2}\\
&=\frac{1}{2}
\end{aligned}
\label{eq9}
\end{equation}

Since $\overline{x}=\overline{y} = 0$, we can derive $\lambda_{1}=\lambda_{2}=\frac{1}{2}$ using \textcolor{blue}{\eqref{eq5}}. $\hfill\square$

\section{Boundary Point Detection Algorithm}\label{sec4}

\subsection{Local Direction Dispersion}
According to Theorem 1, it can be found that the sum of projection variances can reflect the symmetry of KNN distribution on the unit circle.
However, it is insufficient to guarantee high centrality of the query point. For instance, if one half of the KNN are located at (1, 0) and the other half at (-1, 0), we obtain $\lambda_{1}=1$ and $\lambda_{2}=0$ using \textcolor{blue}{\eqref{eq3}}. Although $\lambda_{1}+\lambda_{2}$ reaches the maximum value of 1 and the KNN are symmetrically distributed, they do not fill the unit circle and the centrality is not high. As described in Theorem 2, if the query point has a high centrality, the projection variances of its KNN on all PCs would be equal. The equal of eigenvalues is also known as sphericity of the distribution \coloredcite{r25, r26}. It can measure the homogeneity of different features, thereby indicating the centrality of the query point.

Based on the analyses above, we conclude that the sum and equality of the KNN projection variances together can be used to assess the centrality of the query point. Regarding both factors, we define a centrality metric to measure local direction dispersion (LoDD) for 2-D data
\begin{equation}
L^{(2)} = \omega\left( \lambda_{1}+\lambda_{2} \right)^{2} + 4(1 - \omega)\lambda_{1}\lambda_{2}
\label{eq10}
\end{equation}
where $\lambda_{1}$ and $\lambda_{2}$ are the eigenvalues of the covariance matrix. $\omega$ denotes a regulator that controls the respective effects of the sum and closeness of the two variances, and $0 < \omega < 1$. Based on \textcolor{blue}{\eqref{eq6}}, we have
\begin{equation}
{0 \leq \lambda}_{1}\lambda_{2} \leq \frac{\left( \lambda_{1} + \lambda_{2} \right)^{2}}{4} \leq \frac{1}{4}
\label{eq11}
\end{equation}

Hence, $L^{(2)}$ ranges from 0 to 1, and it can be maximized if and only if $\lambda_{1}=\lambda_{2}=\frac{1}{2}$, while it reaches the minimum value when $\lambda_{1}=\lambda_{2}=0$.

In $d$-dimensional space ($d \geq 2$), we generalize LoDD to
\begin{equation}
L^{(d)} = \omega\left( {\sum\limits_{i = 1}^{d}\lambda_{i}}\right)^{2} + \frac{2d\left(1 - \omega\right)}{d - 1}{\sum\limits_{i,j = 1,i \neq j}^{d}\lambda_{i}}\lambda_{j}
\label{eq12}
\end{equation}

Since
\begin{equation}
{\sum\limits_{i,j = 1,i \neq j}^{d}\lambda_{i}}\lambda_{j} = \frac{1}{2}\left( {\sum\limits_{i = 1}^{d}\lambda_{i}} \right)^{2} - \frac{1}{2}{\sum\limits_{i = 1}^{d}{\lambda_{i}}^{2}}
\label{eq13}
\end{equation}

LoDD is also equivalent to
\begin{equation}
L^{(d)} = \frac{d-\omega}{d-1}\left( {\sum\limits_{i = 1}^{d}\lambda_{i}} \right)^{2}-\frac{d\left(1-\omega\right)}{d-1}{\sum\limits_{i = 1}^{d}{\lambda_{i}}^{2}}
\label{eq14}
\end{equation}

Using Cauchy's inequality, we have 
\begin{equation}
L^{(d)} \leq \frac{d-\omega}{d-1}\left( {\sum\limits_{i = 1}^{d}\lambda_{i}} \right)^{2}-\frac{1-\omega}{d-1}\left( {\sum\limits_{i = 1}^{d}\lambda_{i}} \right)^{2} = \left( {\sum\limits_{i = 1}^{d}\lambda_{i}} \right)^{2}
\label{eq15}
\end{equation}

Meanwhile, \textcolor{blue}{\eqref{eq4}} can be extended to
\begin{equation}
{\sum\limits_{i = 1}^{d}\lambda_{i}}=\frac{1}{k}{\sum\limits_{i = 1}^{d}{\sum\limits_{j = 1}^{k}\left( x_{j}^{(i)} - \overline{x^{(i)}} \right)^{2}}} = 1 - {\sum\limits_{i}{\overline{x^{(i)}}}^{2}} \leq 1
\label{eq16}
\end{equation}
where $x_{j}^{(i)}$ denotes the $i$th coordinate of the $j$th KNN, and $\overline{x^{(i)}}$ represents the $i$th coordinate of the centroid of KNN.

Based on \textcolor{blue}{\eqref{eq15}} and \textcolor{blue}{\eqref{eq16}}, we can obtain
\begin{equation}
0 \leq L^{(d)} \leq 1
\label{eq17}
\end{equation}

When $\lambda_{1}=\lambda_{2}=...=\lambda_{d}=\frac{1}{d}$, $L^{(d)}$ reaches the maximum value of 1; while it gets the minimum if all eigenvalues are equal to 0. Let us take the situation where the KNN are absolutely uniformly distributed on the unit hypersphere of the query point. In this case, the KNN must be symmetrical and centered on the query point, and we have ${\sum_{i = 1}^{d}\lambda_{i}}=1$ using \textcolor{blue}{\eqref{eq16}}. Due to the absolutely uniform distribution, the projection variance of KNN on each PC would be equal, which means that $\lambda_{1}=\lambda_{2}=...=\lambda_{d}=\frac{1}{d}$. The scenario of worst centrality indicates that all KNN coincide and $\lambda_{1}=\lambda_{2}=...=\lambda_{d}=0$.

In the practical implementation, to evade the eigen-decomposition of  covariance matrix $\mathbf{C} \in \mathbb{R}^{d \times d}$, LoDD can be transformed into another form
\begin{equation}
L^{(d)} = \frac{d-\omega}{d-1}tr(\mathbf{C})^{2}-\frac{d(1-\omega)}{d-1}tr(\mathbf{C}^2)
\label{eq18}
\end{equation}

It makes sense since the sum of eigenvalues is equal to the trace of matrix C, and the sum of the squares of these eigenvalues is equal to the trace of matrix $\mathbf{C}^2$. Similar to the Harris corner detector \coloredcite{r45}, which utilizes the trace and determinant to bypass explicit eigenvalue calculations, LoDD also computes the trace to avoid costly matrix decomposition. Such a replacement can streamline operations and enhance computational efficiency.

\subsection{Algorithm Procedure}
Based on the centrality metric LoDD, we propose a boundary point detection algorithm. The algorithm procedure is summarized in Algorithm 1. We first search the KNN $\mathbf{N}(\mathbf{x}_{i})$ of each query point $\mathbf{x}_i$ in the point set $\mathbf{X} \in \mathbb{R}^{d \times n}$ ($n$ refers to the total number of points), and map each KNN onto the unit sphere. Then, we calculate the covariance matrix $\mathbf{C}$ of KNN and LoDD of all points using \textcolor{blue}{\eqref{eq18}}. After arranging all LoDD in ascending order, we identify the points having the first $[100*ratio]\%$ LoDD as boundary points $\mathbf{X}_B$ and detect the remaining as internal points $\mathbf{X}_I$. 

The algorithm has three input parameters, $\omega$, $k$, and $ratio$. Parameter $\omega$ is a regulator that trades off the sum and equality of projection variances. It is fixed at 0.5 for all experiments in this paper. Parameter $k$ is used to determine the number of nearest neighbors, which is commonly relevant to the total number of points, and can be set using an empirical approach \coloredcite{r24}. Parameter $ratio$ determines the proportion of points classified as boundary points. A method for adaptively estimating $ratio$ is developed in Section \ref{sec5}.

\subsection{Time Complexity Analysis}
To assess the computational efficiency, the time complexity of the algorithm is analyzed. Its runtime can be disassembled into three parts
\begin{equation}
T = T_1 + T_2 + T_3
\label{eq19}
\end{equation}
where $T_1$, $T_2$, $T_3$ denote the runtime of KNN search, centrality metric calculation, and LoDD sorting, respectively.

KNN search has a time complexity of $O(T_1)=O(kn\log{n})$. Computing the LoDD consists of KNN mapping and calculation of the covariance matrix. Based on \textcolor{blue}{\eqref{eq18}}, it only considers the diagonal elements of the covariance matrix, so the time complexity is 
\begin{equation}
O(T_2) = O(kn+dn) = O((k+d)n)
\label{eq20}
\end{equation}

The last step divides the point set into boundary points and internal points by arranging the LoDD in ascending order, which has a complexity of $O(T_3)=O(n\log{n})$. In summary, the LoDD algorithm has a overall time complexity of 
\begin{equation}
O(T) = O(kn\log{n}+(k+d)n+n\log{n})=O(kn\log{n}+dn)
\label{eq21}
\end{equation}

The LoDD algorithm embraces parallel computing due to the nature of KNN-based calculation. It can be easily extended to parallel versions using GPGPU and distributed computing techniques such as Apache Spark \coloredcite{r27, r28}.

\begin{table}[t]
\centering
\begin{tabular}{l}
\hline
\textbf{Algorithm 1: LoDD Boundary Point Detector}\\
\hline
\textit{Input:} vertices $\mathbf{X}=\{\mathbf{x}_1,\mathbf{x}_2,...,\mathbf{x}_n\}\in \mathbb{R}^{d \times n}$, $\omega$, $k$, $ratio$ \\
01: \textbf{for} each point $\mathbf{x}_i\in\mathbf{X}$\\
02: ~~~~Search the KNN $\mathbf{N}(\mathbf{x}_{i})$ of point $\mathbf{x}_i$\\
03: ~~~~\textbf{for} each point $\mathbf{x}_j\in\mathbf{N}(\mathbf{x}_{i})$\\
04: ~~~~~~~~Map KNN to the unit sphere by $\mathbf{x}_{j}\leftarrow \frac{( \mathbf{x}_{j} - \mathbf{x}_{i})}{\sqrt{( \mathbf{x}_{j} - \mathbf{x}_{i} )^{T}\left( \mathbf{x}_{j} - \mathbf{x}_{i} \right)}} $\\
05: ~~~~\textbf{end for} \\
06: ~~~~Compute the covariance matrix $\mathbf{C} \in \mathbb{R}^{d \times d}$\\
07: ~~~~Compute LoDD by $L^{(d)}_i \leftarrow \frac{d-\omega}{d-1}tr(\mathbf{C})^{2}-\frac{d(1-\omega)}{d-1}tr(\mathbf{C}^{2})$ \\
08: \textbf{end for}\\
09: Arrange the LoDD as $\tilde{L}^{(d)}_1 \leq \tilde{L}^{(d)}_2 \leq...\leq \tilde{L}^{(d)}_n$ \\
10: \textbf{for} each point $\mathbf{x}_i\in\mathbf{X}$\\
11: ~~~~\textbf{if} $L^{(d)}_i \leq \tilde{L}^{(d)}_{[n \times ratio]}$ \\
12: ~~~~~~~~$\mathbf{X}_B \leftarrow \{\mathbf{x}_i\}$ \\
13: ~~~~\textbf{else} \\
14: ~~~~~~~~$\mathbf{X}_I \leftarrow \{\mathbf{x}_i\}$ \\
15: ~~~~\textbf{end if} \\
16: \textbf{end for}\\
\textit{Output:} $\mathbf{X}_B$ and $\mathbf{X}_I$ \\
\hline
\end{tabular}
\end{table}
\section{Estimation of the Number of Boundary Points}\label{sec5}
As depicted in Algorithm 1, the first $[100*ratio]\%$ points can be identified as boundary points after arranging all LoDD in ascending order. However, parameter $ratio$ is always difficult to specify without enough prior knowledge. In this section, we develop an adaptive method to estimate the number of boundary points based on a grid-structure assumption of data distribution. 

\begin{figure}[t]
\centering
\includegraphics[width=0.94\linewidth]{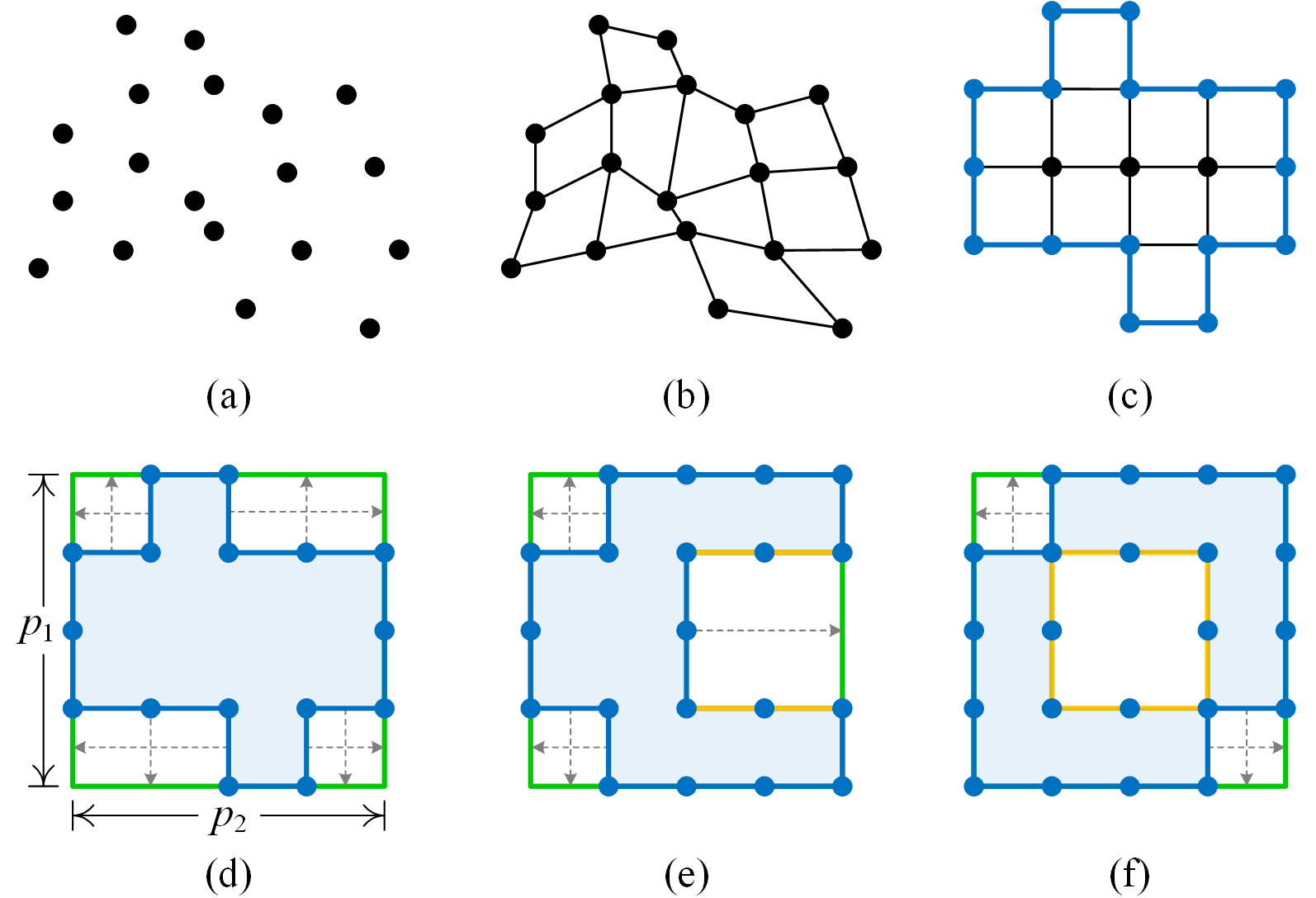}
\caption{Estimation of boundary points based on the grid-structure assumption of data distribution. (a) A 2-D data point set. (b) An irregular grid structure by connecting each point with its neighboring point. Notably, we do not explicitly compute a grid structure, but theoretically estimate the number of boundary points based on this implicit assumption. Nonetheless, we design a grid generation algorithm. It divides the data into multiple equal bins along the X direction to ensure that each bin has similar number of points, then it numbers the points in ascending order of Y for each bin and connects them. Finally, it connects points with the same number along the X direction to generate a grid topological structure. Details of this algorithm can be seen at \url{https://github.com/ZPGuiGroupWhu/lodd/blob/main/lodd_mat/Functions/Intro-GenerateGrid.md}. (c) A regular grid composed of unit square cells, in which the blue points and lines denote the boundary points and periphery of the grid, respectively. (d)-(f) Projecting the non-overlapping outer edges of the grid cells onto the MBR orthogonally, where the blue areas represent the grid extent, the green and yellow segments denote the orthogonal projections on MBR, and the inner segments of a concave or ring-shaped grid, respectively.}
\label{fig4}
\end{figure}

\subsection{Upper and Lower Bounds}
To estimate the upper and lower bounds of the number of boundary points, we introduce a latent grid topological structure to represent the 2-D data distribution. We provide an example in Fig. \ref{fig4}\textcolor{blue}{(a)}, where a bunch of points can generate an irregular grid by connecting each point with its neighboring points, as shown in Fig. \ref{fig4}\textcolor{blue}{(b)}. We can regularize the structure into a grid composed of unit square cells. In this case, the boundary points are the ones on the periphery of the grid extent, which are denoted as blue points in Fig. \ref{fig4}\textcolor{blue}{(c)}, and their number is equal to the periphery length. Actually, a convex grid has the same perimeter as the corresponding Minimum Bounding Rectangle (MBR), since the non-overlapping outer edges of the cells can be projected orthogonally onto the MBR (see the green segments in Fig. \ref{fig4}\textcolor{blue}{(d)}). Whereas, a concave or ring-shaped grid has a larger perimeter than MBR due to the length of inner segments (see the yellow segments in Fig. \ref{fig4}\textcolor{blue}{(e)} and \textcolor{blue}{(f)}). Therefore, the number of boundary points $B$ is equal to the MBR perimeter plus the length of inner line segments for 2-D data
\begin{equation}
B=2(p_1+p_2-2)+l
\label{eq22}
\end{equation}
where $p_1$ and $p_2$ represent the maximum number of rows and columns of points, respectively, and $l$ denotes the length of inner line segments. 

Based on the grid-structure assumption, there are at least two rows or columns of points, and the number of potential points fully-arranged in the MBR with $p_1$ rows and $p_2$ columns is not less than the total number of points $n$. Suppose that $p_2\leq p_1$, thus we can define a set of constraints
\begin{equation}
R( p_{1},{p}_{2} ) = \left\{ \begin{matrix}
{2 \leq p_{2} \leq p_{1} \leq \frac{n}{2}} \\
~ \\
 n \leq {p_{1}p_{2}}
\end{matrix} \right.
\label{eq23}
\end{equation}

This set of constraints yields the upper bound
\begin{equation}
{\underset{R(p_{1},{p}_{2})}{\mathit{\sup}}B} = n
\label{eq24}
\end{equation}

When $p_1=\frac{n}{2}$, $p_2=2$, and $l=0$, $B$ is equal to the total number of points $n$. In this scenario, all points are the boundary points, and the square cells of the grid are arranged in a column.

Meanwhile, as depicted in \textcolor{blue}{\eqref{eq25}}, the lower bound of $B$ follows the AM-GM inequality when $p_1=p_2=\sqrt{n}$. In this case, the grid is a square with the same number of rows and columns.
\begin{equation}
{\underset{R(p_{1},{p}_{2})}{\mathit{\inf}}B} = 4\sqrt{n}-4
\label{eq25}
\end{equation}

The grid-structure assumption is feasible in 2-D space, since the latent grid structure can depict point patterns of any shape and is not affected by the density heterogeneity.

\begin{figure}[t]
\centering
\includegraphics[width=0.94\linewidth]{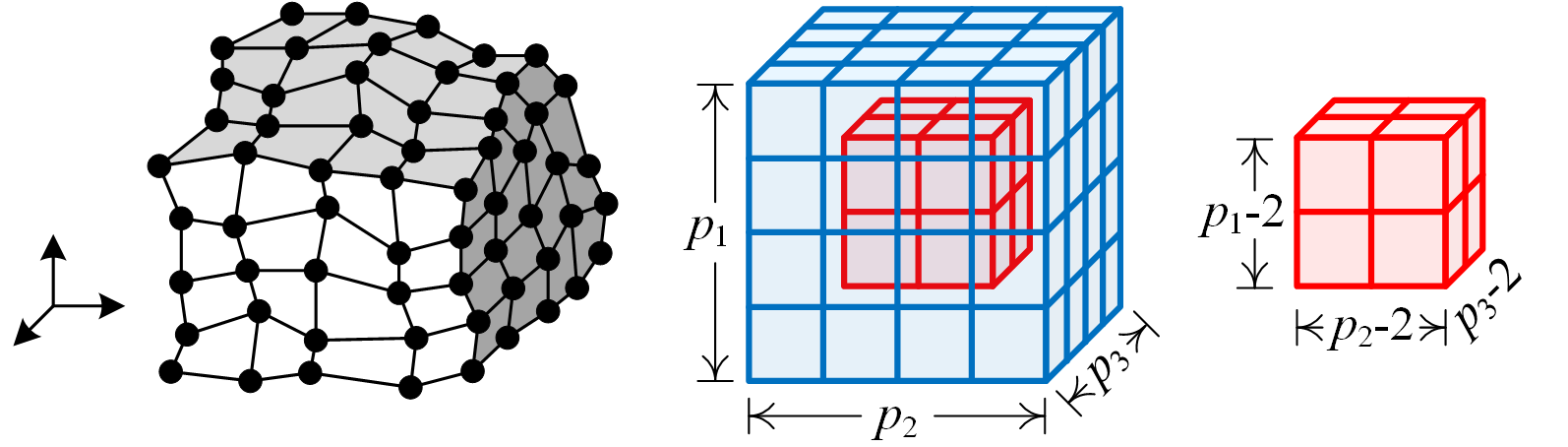}
\caption{A 3-D example for estimating the number of boundary points in high-dimensional space, where the boundary points lie on the blue surface and the internal points are in the red hypercuboid.}
\label{fig5}
\end{figure}

To estimate the upper and lower bounds of $B$ in higher-dimensional space ($d > 2$), we make a more rigorous assumption that the data can form an irregular hypercuboid with points fully-arranged in $p_i$ rows in the $i$th dimension implicitly (see Fig. \ref{fig5}). It can be found that the boundary points lie on the outer surface of the hypercuboid. Suppose $p_d\leq...\leq p_2 \leq p_1$, then the constraint is
\begin{equation}
R( p_{1},{p}_{2},...,p_{d} ) = \left\{ \begin{matrix}
{ 2 \leq p_d \leq ... \leq p_2 \leq p_1 \leq \frac{n}{2^{d - 1}}} \\
~ \\
{p_1p_2...p_{d} = n}
\end{matrix} \right.
\label{eq26}
\end{equation}

$B$ reaches the same upper bound $n$ as in the 2-D space. It occurs when $p_1={n}/{2^{d - 1}}$ and $p_2=p_3=...=p_d=2$, which means that the data hypercuboid is formed by a row of hypercubes arranged in sequence. 

As illustrated in Fig. \ref{fig5}, the number of internal points is equal to $\prod_{i = 1}^{d}( p_{i} - 2)$, so the objective to calculate the lower bound can be formulated
\begin{equation}
\arg\max{\prod\limits_{i = 1}^{d}{( p_{i} - 2),}}~s.t.~p_{1}p_{2}\ldots p_{d} = n
\label{eq27}
\end{equation}

Using Lagrange multiplier, we construct a function
\begin{equation}
\psi = {\prod\limits_{i = 1}^{d}( p_{i} - 2)} - \xi( p_{1}p_{2}\ldots p_{d} - n)
\label{eq28}
\end{equation}

Then, we can solve the partial derivatives and set them to zero. The system of equations we need to solve is
\begin{equation}
\left\{ \begin{matrix}
{\frac{\partial\psi}{\partial\xi} = n - p_{1}p_{2}\ldots p_{d} = 0} \\
~\\
{\frac{\partial\psi}{\partial p_{i}} = {\prod\limits_{j \neq i}( p_{j} - 2 )} - \xi{\prod\limits_{j \neq i}p_{j}} = 0}
\end{matrix} \right.
\label{eq29}
\end{equation}

By solving \textcolor{blue}{\eqref{eq29}}, we have
\begin{equation}
p_1= p_2 =...= p_d = \sqrt[d]{n}
\label{eq30}
\end{equation}

Thus, we can obtain the lower bound
\begin{equation}
{\inf\limits_{R(p_{1},\mathit{p}_{2},...,p_{d})}B} = n - \left( \sqrt[d]{n} - 2 \right)^{d},~s.t.~d \leq \log_2n
\label{eq31}
\end{equation}

In summary, we can estimate an approximate range of $B$ based on the grid-structure assumption of data distribution
\begin{equation}
n - ( \sqrt[d]{n} - 2 )^{d} \leq B \leq n,~s.t.~d \leq \log_2n
\label{eq32}
\end{equation}

\begin{figure*}[t]
\centering
\includegraphics[width=0.9\linewidth]{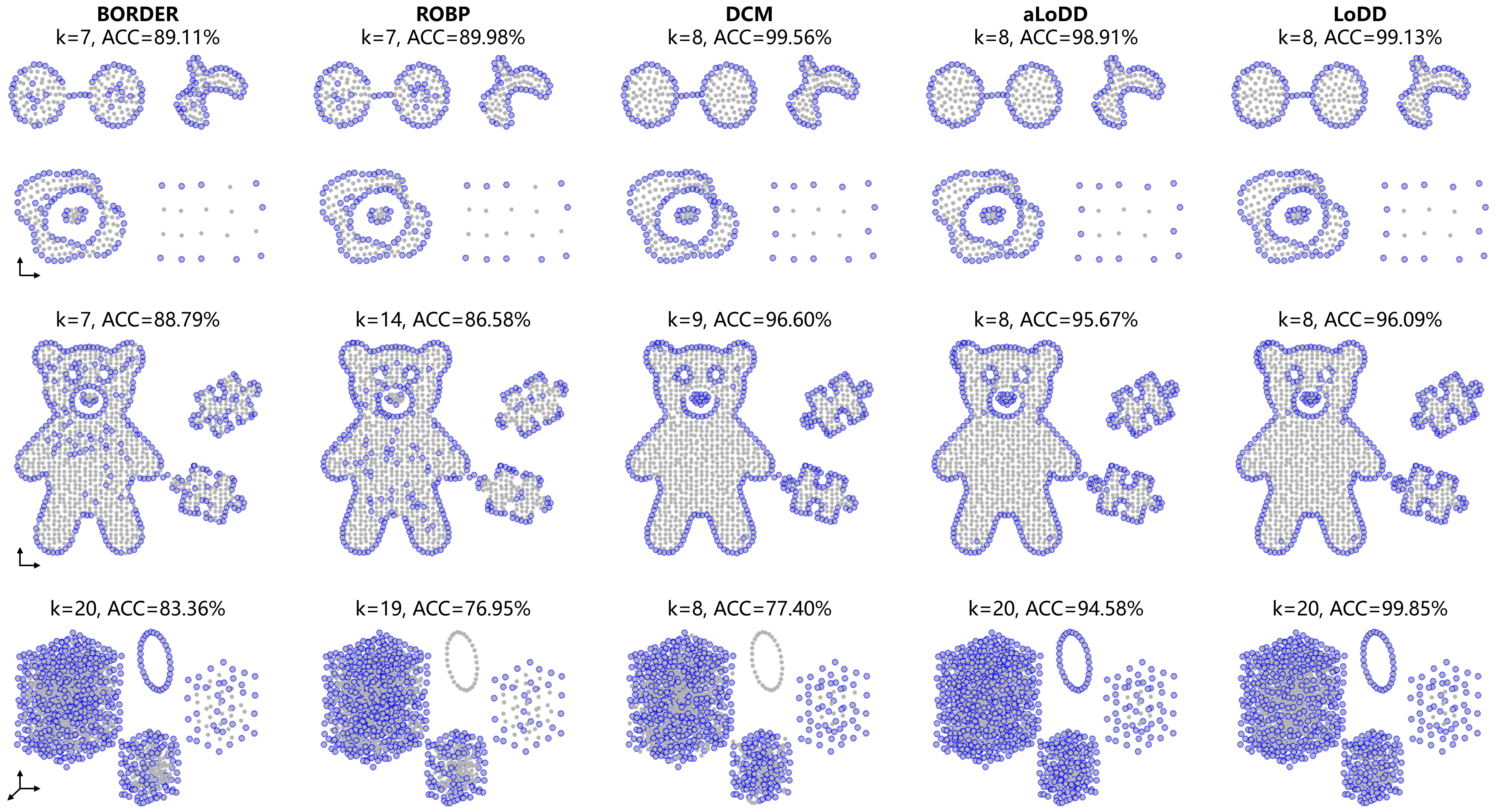}
\caption{Boundary detection results of the BORDER, ROBP, DCM, and LoDD with adaptive and optimal $ratio$ algorithms on DS1-DS3, where the boundary points are represented by blue circles.}
\label{fig6}
\end{figure*}

\subsection{Determination of Ratio}
The manifold hypothesis posits that high-dimensional data in the real world always lie along low-dimensional latent manifolds and the intrinsic dimension is lower than the feature dimension \coloredcite{r29}. From a centrality perspective, all points in the manifold would be regarded as boundary points in a higher-dimensional space, which can also be proved by \textcolor{blue}{\eqref{eq32}} that the number of points $B$ approaches $n$ as $d$ approaches infinity. It means that adopting the feature dimension $d$ may lead to an overestimation of $B$ for high-dimensional data. To address this problem, we determine the intrinsic dimension by leveraging PCA to manipulate the data and selecting the dimension $D$ that can make the cumulative contribution rate slightly above 0.8. In practice, to avoid generating an excessively large $B$, we set its upper limit to $0.5n$. Therefore, we can estimate the number of boundary points in a single cluster to be

\begin{equation}
B=\min\left(0.5n,~n - \left( \sqrt[D]{n} - 2 \right)^{D}\right)
\label{eq33}
\end{equation}

However, there is usually more than one cluster in the data. We need to obtain an estimate of the number of points in each cluster before determining $ratio$. If the exact number of clusters $c$ is known in advance, we can roughly regard each cluster to have the same number of points. In this case, the total number of boundary points is equal to the number of boundary points in a single cluster with $n/c$ points multiplied by $c$. Using \textcolor{blue}{\eqref{eq33}}, $ratio$ can be determined

\begin{equation}
\begin{aligned}
ratio = \frac{B}{n} &=\frac{\min\left(0.5n,~c\left(\frac{n}{c} - \left( \sqrt[D]{\frac{n}{c}} - 2 \right)^{D}\right)\right)}{n}\\
&=\min\left(0.5,~1-\frac{\left( \sqrt[D]{n} - 2\sqrt[D]{c}\right)^{D}}{n}\right) 
\label{eq34}
\end{aligned}
\end{equation}

If $c$ is unknown, we can construct a KNN graph and regard each connected component as a cluster. The result of the KNN graph can also be used in the LoDD algorithm, which does not introduce an excessive extra time cost. Suppose the KNN graph consists of $m$ connected components, with $n_1,n_2,...,n_m$ points, respectively, then we can determine the ratio as

\begin{equation}
ratio = \frac{B}{n} = \min\left(0.5,~1-\frac{\sum_{i = 1}^{m}\left( \sqrt[D]{n_{i}} - 2\right)^{D}}{n}\right)
\label{eq35}
\end{equation}
  
\begin{table}[t]
\centering
\caption{Information of 12 Real-world Benchmarks}
\begin{tabular}{c c c c c}
\hline
\textbf{Dataset} & \textbf{Description} & \textbf{\#Instances} & \textbf{\#Features} & \textbf{\#Classes}\\
\hline
Breast-Cancer & UCI & 683 & 10 & 2 \\
PenDigits & UCI & 10,992 & 16 & 10 \\
Control & UCI & 600 & 60 & 6 \\
Mice & UCI & 1,077 & 77 & 8\\
Digits &  Digit Images & 5,620 & 8×8 & 10 \\
MNIST10k & Digit Images & 10,000 & 28×28 & 10 \\
Yale & Face Images & 165 & 64×64 & 15 \\
FEI-1 & Face Images & 700 & 480×640 & 50 \\
TDT2-5 & Text Documents  & 2,173 & 36,771 & 5 \\
TDT2-10 & Text Documents  & 4,315 & 36,771 & 10 \\
Segerstolpe & scRNA-seq & 2,133 & 22,757 & 13 \\
Xin & scRNA-seq  & 1,449 & 33,889 & 4 \\
Levine & CyTOF  & 265,627 & 32 & 14 \\
Samusik & CyTOF  & 841,644 & 40 & 24\\
\hline
\end{tabular}
\label{tab1}
\end{table}

\section{Experiments}\label{sec6}
\subsection{Performance on Synthetic Datasets}
To validate the effectiveness, we compared the LoDD algorithm with three baselines, including two typical density-based methods, BORDER \coloredcite{r14} and ROBP \coloredcite{r17}, and a geometry-based detector DCM \coloredcite{r24}, on three synthetic datasets (DS1-DS3). Since all methods are based on KNN, we varied $k$ in the same range from 5 to 20 with an interval of 1 and specified $ratio$ as the true proportion of boundary points in each dataset. We also performed LoDD using the $ratio$ adaptively obtained from \textcolor{blue}{\eqref{eq34}} and denoted it as aLoDD. The ACCuracy (ACC) metric was utilized to quantitatively evaluate the results.

\begin{table*}
\centering
\caption{The Highest ACC (Growth) (\%) of the K-means Algorithm Jointly with Different Boundary Point Detectors on 14 Real-world Datasets (The Best Results are Highlighted in Bold)}
\begin{tabular}{c c c c c c c c c}
\hline
Dataset & KM & KM+BORDER & KM+ROBP & KM+NC & KM+DCM & KM+LDIV & KM+aLoDD & KM+LoDD\\
\hline
Breast-Cancer & 96.19 & 96.19 (+0.00) & 96.34 (+0.15) & 95.90 (-0.29) & 97.07 (+0.88) & \textbf{97.22 (+1.03)} & 96.34 (+0.15) & \textbf{97.22 (+1.03)}\\
PenDigits & 67.90 & 77.24 (+9.34) & 77.30 (+9.40) & 76.97 (+9.07) & 76.98 (+9.08) & 68.40 (+0.50) & 68.51 (+0.61) & \textbf{77.38 (+9.48)}\\
Control & 56.83 & 57.50 (+0.67) & 57.33 (+0.50) & 67.67 (+10.84) & 70.83 (+14.00) & 70.67 (+13.84) & 70.33 (+13.50) & \textbf{72.67 (+15.84)}\\
Mice & 34.73 & 36.55 (+1.82) & 35.81 (+1.08) & 35.65 (+0.92) & 37.79 (+3.06) & 39.20 (+4.47) & \textbf{39.55 (+4.82)} & \textbf{39.55 (+4.82)}\\
Digits & 74.79 & 76.81 (+2.02) & 78.72 (+3.93) & 77.62 (+2.83) & \textbf{82.99 (+8.20)} & 82.74 (+7.95) & 81.65 (+6.86) & 81.65 (+6.86)\\
MNIST10k & 52.90 & 53.90 (+1.00) & 53.75 (+0.85) & 58.69 (+5.79) & 56.45 (+3.55) & 61.39 (+8.49) & \textbf{62.13 (+9.23)} & \textbf{62.13 (+9.23)}\\
Yale & 66.26 & 76.07 (+9.81) & 76.07 (+9.81) & 76.07 (+9.81) & 68.71 (+2.45) & 71.17 (+4.91) & \textbf{77.30 (+11.04)} & \textbf{77.30 (+11.04)}\\
FEI-1 & 50.86 & 54.43 (+3.57) & 57.86 (+7.00) & 58.00 (+7.14) & 55.57 (+4.71) & 58.86 (+8.00) & 58.29 (+7.43) & \textbf{59.57 (+8.71)}\\
TDT2-5 & 39.58 & 49.42 (+9.84) & 49.06 (+9.48) & 52.37 (+12.79) & 43.86 (+4.28) & 49.79 (+10.21) & \textbf{57.16 (+17.58)} & \textbf{57.16 (+17.58)}\\
TDT2-10 & 49.32 & 50.95 (+1.63) & 51.99 (+2.67) & 50.35 (+1.03) & 53.25 (+3.93) & 51.95 (+2.63) & \textbf{54.79 (+5.47)} & \textbf{54.79 (+5.47)}\\
Segerstolpe & 63.67 & 64.23 (+0.56) & 63.99 (+0.32) & 63.95 (+0.28) & \textbf{79.00 (+15.33)} & 78.43 (+14.76) & 72.62 (+8.95) & 74.78 (+11.11)\\
Xin & 79.30 & 81.16 (+1.86) & 81.92 (+2.26) & 81.30 (+2.00) & 81.65 (+2.35) & 77.09 (-2.21) & 82.88 (+3.58) & \textbf{86.34 (+7.04)}\\
Levine & 71.63 & 70.22 (-1.41) & 71.87 (+0.24) & 69.81 (-1.82) & 76.03 (+4.40) & 71.15 (-0.48) & 76.76 (+5.13) & \textbf{77.53 (+5.90)}\\
Samusik & 55.33 & 56.23 (+0.90) & 56.71 (+1.38) & 55.39 (+0.06) & 56.27 (+0.94) & 57.73 (+2.40) & 62.08 (+6.75) & \textbf{63.16 (+7.83)}\\
\hdashline
Ave. Growth & / & +2.97 & +3.53 & +4.32 & +5.51 & +5.46 & +7.22 & \textbf{+8.71}\\
\hline
\end{tabular}
\label{tab2}
\end{table*}

\begin{table*}
\centering
\caption{The Highest NMI (Growth) (\%) of the K-means Algorithm Jointly with Different Boundary Point Detectors on 14 Real-world Datasets (The Best Results are Highlighted in Bold)}
\begin{tabular}{c c c c c c c c c}
\hline
Dataset & KM & KM+BORDER & KM+ROBP & KM+NC & KM+DCM & KM+LDIV & KM+aLoDD & KM+LoDD\\
\hline
Breast-Cancer & 75.46 & 75.37 (-0.09) & 76.02 (+0.56) & 74.78 (-0.68) & 79.92 (+4.46) & \textbf{80.60 (+5.14)} & 76.07 (+0.61) & \textbf{80.60 (+5.14)}\\
PenDigits & 67.16 & 68.79 (+1.63) & 69.36 (+2.20) & 68.25 (+1.09) & 69.79 (+2.63) & 69.51 (+2.35) & 69.24 (+2.08) & \textbf{70.58 (+3.42)}\\
Control & 75.42 & 76.77 (+1.35) & 76.78 (+1.36) & 78.28 (+2.86) & 78.43 (+3.01) & 80.71 (+5.29) & 80.42 (+5.00) & \textbf{80.87 (+5.45)}\\
Mice & 35.14 & 38.70 (+3.56) & 38.65 (+3.51) & 37.01 (+1.87) & 39.99 (+4.85) & 42.12 (+6.98) & 42.96 (+7.82) & \textbf{43.48 (+8.34)}\\
Digits & 72.91 & 76.22 (+3.31) & 77.60 (+4.69) & 76.00 (+3.09) & \textbf{80.77 (+7.86)} & 78.54 (+5.63) & 77.62 (+4.71) & 77.62 (+4.71)\\
MNIST10k & 47.19 & 51.02 (+3.83) & 51.27 (+4.08) & 52.04 (+4.85) & 51.14 (+3.95) & 54.34 (+7.15) & \textbf{55.44 (+8.25)} & \textbf{55.44 (+8.25)}\\
Yale & 70.88 & 77.67 (+6.79) & 78.43 (+7.55) & 77.67 (+6.79) & 72.03 (+1.15) & 74.53 (+3.65) & 77.97 (+7.09) & \textbf{78.53 (+7.65)}\\
FEI-1 & 78.58 & 83.07 (+4.49) & 83.29 (+4.71) & 82.02 (+3.44) & 79.91 (+1.33) & 83.45 (+4.87) & 82.01 (+3.43) & \textbf{84.01 (+5.43)}\\
TDT2-5 & 42.15 & 48.66 (+6.51) & 50.40 (+8.25) & 49.34 (+7.19) & 48.24 (+6.09) & 50.08 (+7.93) & 50.57 (+8.42) & \textbf{50.62 (+8.47)}\\
TDT2-10 & 62.65 & 63.12 (+0.47) & 64.07 (+1.42) & 63.53 (+0.88) & \textbf{71.13 (+8.48)} & 67.79 (+5.14) & 65.44 (+2.79) & 68.43 (+5.78)\\
Segerstolpe & 76.19 & 76.62 (+0.43) & 76.58 (+0.39) & 76.94 (+0.75) & 81.95 (+5.76) & \textbf{84.20 (+8.01)} & 79.35 (+3.16) & 82.44 (+6.25)\\
Xin & 70.74 & 75.51 (+4.77) & 76.11 (+5.37) & 76.30 (+5.56) & 75.85 (+5.11) & 74.02 (+3.28) & 73.87 (+3.13) & \textbf{76.49 (+5.57)}\\
Levine & 79.20 & 79.68 (+0.48) & 80.49 (+1.29) & 81.10 (+1.90) & 81.47 (+2.37) & 80.40 (+1.20) & 84.31 (+5.11) & \textbf{84.34 (+5.14)}\\
Samusik & 74.16 & 75.25 (+1.09) & 73.89 (-0.27) & 75.21 (+1.05) & 76.47 (+2.31) & 75.18 (+1.02) & 80.09 (+5.93) & \textbf{80.58 (+6.42)}\\
\hdashline
Ave. Growth & / & +2.76 & +3.22 & +2.90 & +3.80 & +4.62 & +4.82 & \textbf{+6.16}\\
\hline
\end{tabular}
\label{tab3}
\end{table*}

The optimal results in the parameter space are presented in Fig. \ref{fig6}. DS1 contains two weakly-connected spherical clusters, a non-spherical dense cluster, and a sparse cluster. Meanwhile, a ring cluster surrounds a spherical cluster, appearing as an island distribution. The two density-based methods, BORDER and ROBP cannot identify the boundary points of the sparse cluster due to the density heterogeneity. However, LoDD and aLoDD achieved similar performance to DCM and only misidentified one and two more points than it. Both DCM and LoDD algorithms can cope with weak connectivity and density heterogeneity. DS2 includes a bear-shaped cluster and two puzzle-shaped clusters. Both BORDER and ROBP misidentified some points inside the bear body as boundary points and failed to find the boundary points in the concave areas. In contrast, DCM, aLoDD, and LoDD yielded higher accuracy. With respect to the 3-D dataset DS3, ROBP and DCM are unable to detect the boundary points of the ring, while LoDD exhibited a distinct advantage to handle higher-dimensional manifolds.

\subsection{Performance in Cluster Analysis}
Boundary peeling has a positive impact on clustering quality \coloredcite{r30}, in turn, the clustering accuracy can implicitly reflect the effectiveness of boundary point detectors. To verify the performance on high-dimensional data, we utilized the boundary point detectors to identify the boundary points and peel them out. Then, we conducted the K-means (KM) algorithm to cluster the internal points, and assigned the cluster label of the nearest internal point to each boundary point to obtain the final clustering results. Five typical boundary point detectors, BORDER \coloredcite{r14}, ROBP \coloredcite{r17}, NC \coloredcite{r22}, DCM \coloredcite{r24}, and LDIV \coloredcite{r49} were selected for comparison. Since DCM is extremely computing-intensive in high-dimensional space, we first applied PCA to reduce the original data to a 2-D space, and then input this reduced data into DCM for boundary point identification.

To determine the initial cluster centers for K-means, we followed the idea of the density peak clustering algorithm, which suggests that each cluster center tends to have a high local density and is far from other cluster centers \coloredcite{r31}. We defined the local $density$ as the number of points within a cutoff distance. The cutoff distance is set to be the 5$\%$ shortest distance among all point-pair distances. Let $minDis$ denote the minimum distance between the current point and any other point with higher density. We normalized the $density$ and $minDis$ to [0, 1] and calculated a $score$
\begin{equation}
score=density+minDis
\label{eq36}
\end{equation}

\begin{figure}[t]
\centering
\includegraphics[width=0.98\linewidth]{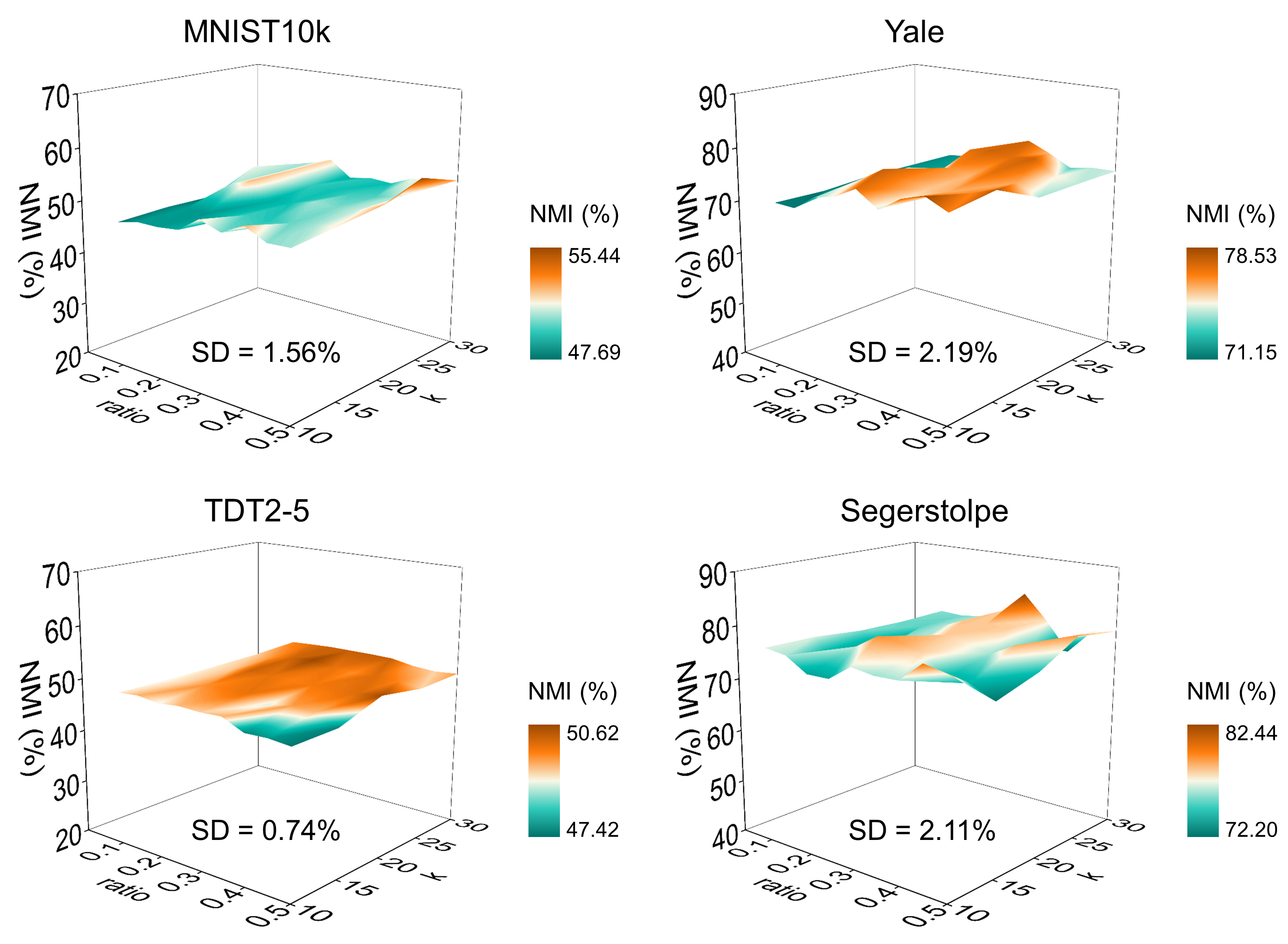}
\caption{Parameter sensitivity analysis of $k$ and $ratio$ by performing KM+LoDD on four real-world datasets.}
\label{fig7}
\end{figure}

If $c$ denotes the actual number of clusters, then points with the $c$ largest score values were selected as the initial cluster centers for K-means. In terms of the parameter settings for the boundary point detectors, we specified $k$ and $ratio$ from 10 to 30 with an interval of 5, and from 0.1 to 0.5 with an interval of 0.05, respectively. With respect to aLoDD, the parameter $ratio$ was automatically determined according to \textcolor{blue}{\eqref{eq34}}.

\textbf{Datasets and preprocessing:} Table \ref{tab1} presents the information of 12 real-world datasets of different data types. Breast-Cancer, PenDigits, Control, and Mice are four UCI benchmarks \coloredcite{r32}. Digits and MNIST10k \coloredcite{r33} are two typical handwritten digit images with size of 8×8 and 28×28 respectively. We converted them from 2-D image matrices to 1-D vectors. Yale contains 165 face images with size of 64×64 from 15 distinct individuals \coloredcite{r34}, while FEI-1 is composed of 700 face images taken from 50 individuals and the images are in size of 480×640 pixels \coloredcite{r35}. We extracted the Gabor, Gist and LBP (Local Binary Pattern) features from the face images. PCA was used to individually reduce each feature to 50 dimensions, and then these three features were concatenated into a 150-D vector as input. TDT2-5 and TDT2-10 are two subsets of the original TDT2 document dataset \coloredcite{r39}, which contain 2,173 and 4,315 documents with 36,771 words, respectively. We obtained the Term Frequency-Inverse Document Frequency (TF-IDF) \coloredcite{r40} of each word, and embedded the TF-IDF features into a 100-D vector. Segerstolpe \coloredcite{r36} and Xin \coloredcite{r37} are two scRNA-seq (single-cell RNA sequencing) datasets. They respectively consist of 2,133 and 1,449 cells, corresponding to 22,757 and 33,889 genes. We utilized Seurat software \coloredcite{r38} to conduct the standard preprocessing pipeline for scRNA-seq data, including quality control, normalization, selecting the top 2,000 highly variable genes (HVG), scaling, and embedding the HVG into a 50-D vector by PCA. Levine \coloredcite{r46} and Samusik \coloredcite{r47} are two mass cytometry (CyTOF) datasets, containing 265,627 and 841,644 bone marrow cells with 32 and 40 surface markers, respectively. We followed the standard preprocessing steps in \coloredcite{r48} by applying an arcsin transformation with a standard cofactor of 5. We adopted min-max normalization to preprocess all datasets by scaling the features to [0, 1] before clustering.

\begin{figure}[t]
\centering
\includegraphics[width=0.85\linewidth]{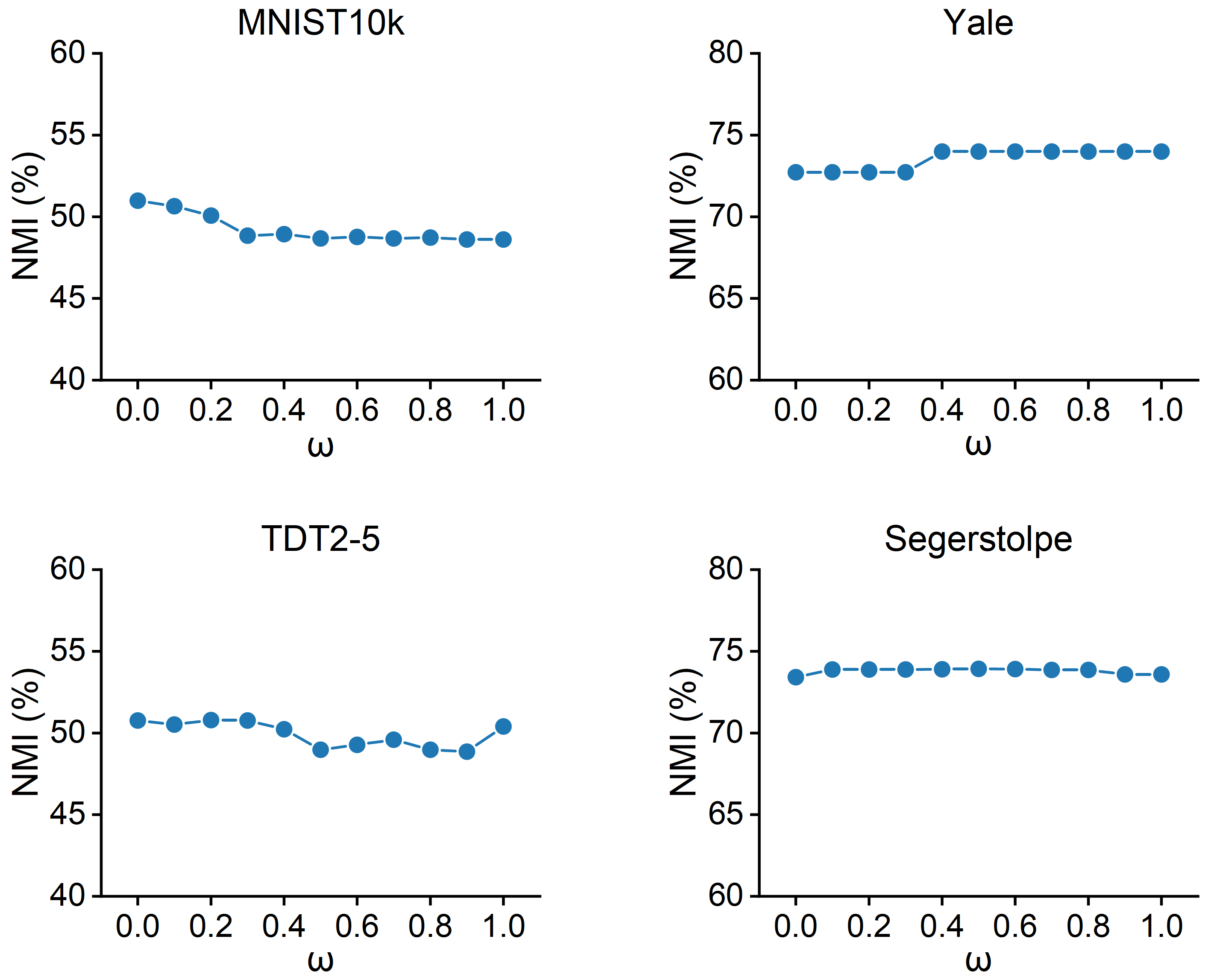}
\caption{The impact of the regulator coefficient $\omega$ on the clustering performance of KM+LoDD.}
\label{fig8}
\end{figure}

\textbf{Evaluation metrics:} To quantitatively evaluate the clustering performance, we adopted two widely used evaluation metrics, ACC and NMI (Normalized Mutual Information). ACC refers to the accuracy rate of the predicted results compared to the true labels. Given the true label vector $l=(l_1,l_2,\ldots,l_n)\in\mathbb{R}^n$ and the predicted label vector $r=(r_1,r_2,\ldots,r_n)\in\mathbb{R}^n$, ACC can be defined as
\begin{equation}
ACC = \frac{\sum\limits_{i = 1}^{n}{\delta\left( {l_{i},map\left( r_{i} \right)} \right)}}{n}
\label{eq37}
\end{equation}
where $\delta(\cdot)$ denotes an indicator function

\begin{equation}
\delta\left( {x,y} \right) = \left\{ \begin{matrix}
{1~~if~~x = y} \\
{~0~~otherwise}
\end{matrix} \right.
\label{eq38}
\end{equation}
$map(\cdot)$ is a mapping function that maps each predicted label to one of the true cluster labels. Commonly, the best mapping can be found using the Kuhn-Munkres or Hungarian algorithms.

NMI measures the agreement of predicted and true assignments by ignoring permutations. Its formulation is
\begin{equation}
NMI = \frac{\sum\limits_{i = 1}^{|S|}{{\sum\limits_{j = 1}^{|Q|}\left| {S_{i} \cap Q_{j}} \right|}\log\frac{n\left| {S_{i} \cap Q_{j}} \right|}{\left| S_{i} \right|\left| Q_{j} \right|}}}{\sqrt{\left( {{\sum\limits_{i = 1}^{|S|}\left| S_{i} \right|}{\log\frac{\left| S_{i} \right|}{n}}} \right)\left( {\sum\limits_{j = 1}^{|Q|}{\left| Q_{j} \right|{\log\frac{\left| Q_{j} \right|}{n}}}} \right)}}
\label{eq39}
\end{equation}
where $S_i$ denotes the point set in the $i$th predicted cluster, while $Q_j$ denotes that in the $j$th cluster of the true labels.

\textbf{Results:} The highest ACC and NMI scores in the parameter space are illustrated in Tables \ref{tab2} and \ref{tab3}. Overall, boundary point detection can bolster the clustering performance of the K-means algorithm, with LoDD achieving the most significant accuracy enhancement. On average, it increased ACC by 8.71\% and NMI by 6.16\%, and made improvements of over 5\% in ACC and NMI across 12 datasets. When equipped with the adaptive $ratio$, LoDD yielded the best ACC on five datasets and improved ACC by an average of 7.22\% and NMI by 4.82\%. DCM achieved promising results as well as the recently proposed LDIV method. They obtained the highest ACC scores on two and one datasets, respectively. Nonetheless, LoDD led DCM and LDIV by 3.20\% and 3.25\% in terms of average ACC growth rate.

\begin{figure}[t]
\centering
\includegraphics[width=1\linewidth]{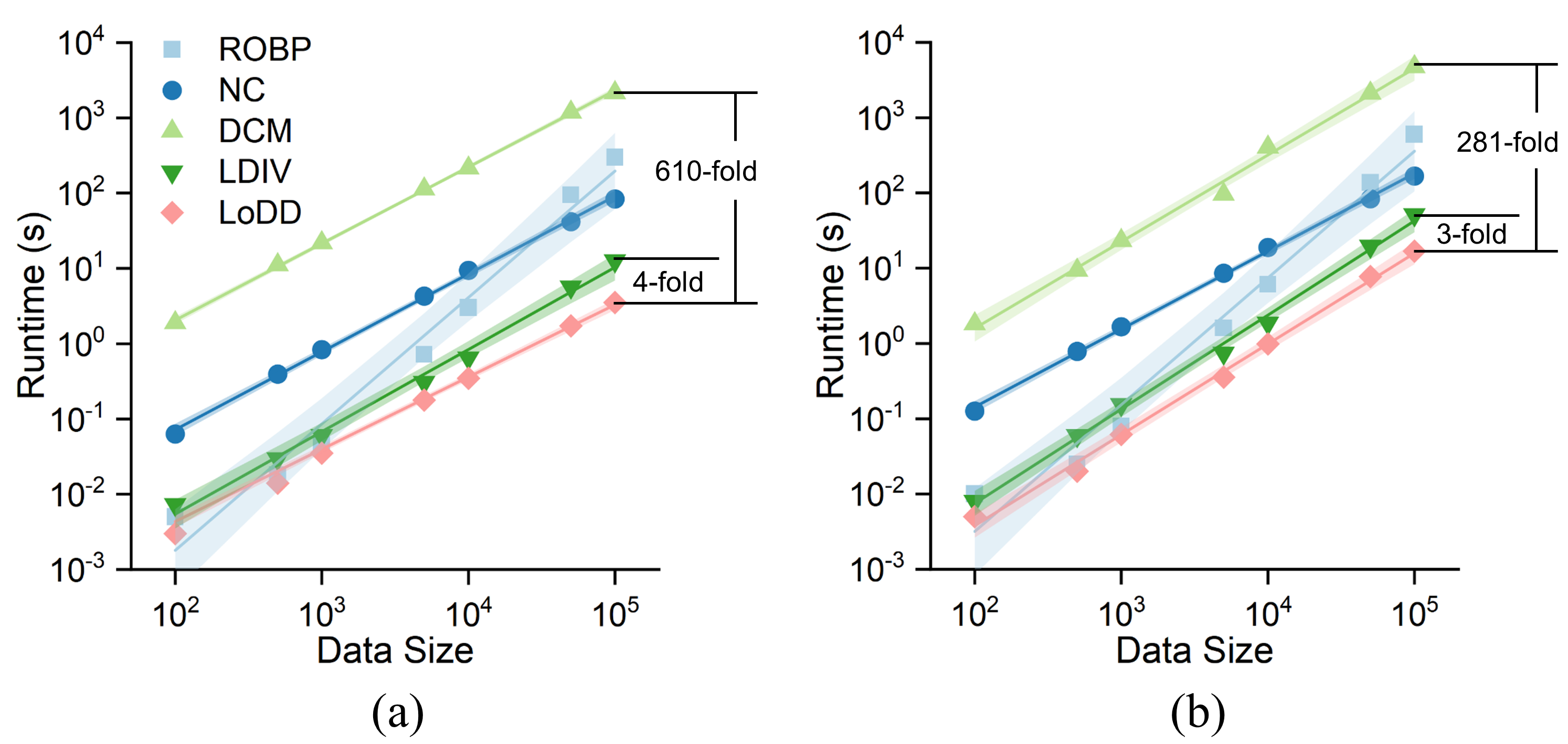}
\caption{Runtimes of five boundary point detectors implemented in (a) MATLAB and (b) Python across the simulated 10-D datasets with different sizes. The shading denotes a 95\% confidence band.}
\label{fig9}
\end{figure}

\subsection{Parameter Sensitivity Analysis} 
The algorithm has three parameters, $k$, $ratio$, and $\omega$. To assess the sensitivity of $k$ and $ratio$, we performed KM+LoDD with $k$ from 10 to 30 and $ratio$ from 0.1 to 0.5 on four real-world datasets of different data types. As shown in Fig. \ref{fig7}, the clustering performance is relatively stable in the parameter space with a low standard deviation (SD) of less than 3\%. It demonstrates that cluster task is insensitive to $k$ and $ratio$ of LoDD. Meanwhile, we varied the regulator $\omega$ from 0 to 1 and evaluated the clustering accuracy of KM+LoDD in Fig. \ref{fig8}. The NMI metric presents a stable trend that illustrates the robustness of LoDD to $\omega$. Based on our substantial empirical experiments, $10\sim30$ and $0.2\sim0.5$ are the suggested default ranges of $k$ and $ratio$ for handling datasets with sample sizes ranging from hundreds to millions, respectively, and the regulator $\omega$ is recommended as 0.5.

\subsection{Time Efficiency Analysis} 
The time efficiency analysis was conducted on a commodity desktop computer with a 12-core Intel i7 processor and 128 GB RAM. We selected four mainstream boundary point detectors for comparison and implemented them in MATLAB and Python. The average runtimes were measured using a varying $k$ from 10 to 30 with an interval of 5 and a fixed $ratio=0.1$. Fig. \ref{fig9} shows the runtime comparison on 10-D simulated datasets with different number of points. As MATLAB uses just-in-time compilation to accelerate the code, MATLAB versions of these algorithms are generally more efficient than the Python versions, but the overall trend remains similar. DCM has low computational efficiency when handling high-dimensional data, with the longest running time at this data scale. The runtime of ROBP escalates most rapidly as the dataset expands. By contrast, LoDD exhibits the best scaling performance with increasing data size. Its MATLAB version runs in around 3.5 seconds on 100k points, which is 610, 85, 24, and 4 times faster than DCM, ROBP, NC, and LDIV, while the runtime of its Python version is 16.8 seconds, which is 281, 36, 10, and 3 times less than these four baselines, respectively.

\begin{figure}[t]
\centering
\includegraphics[width=0.98\linewidth]{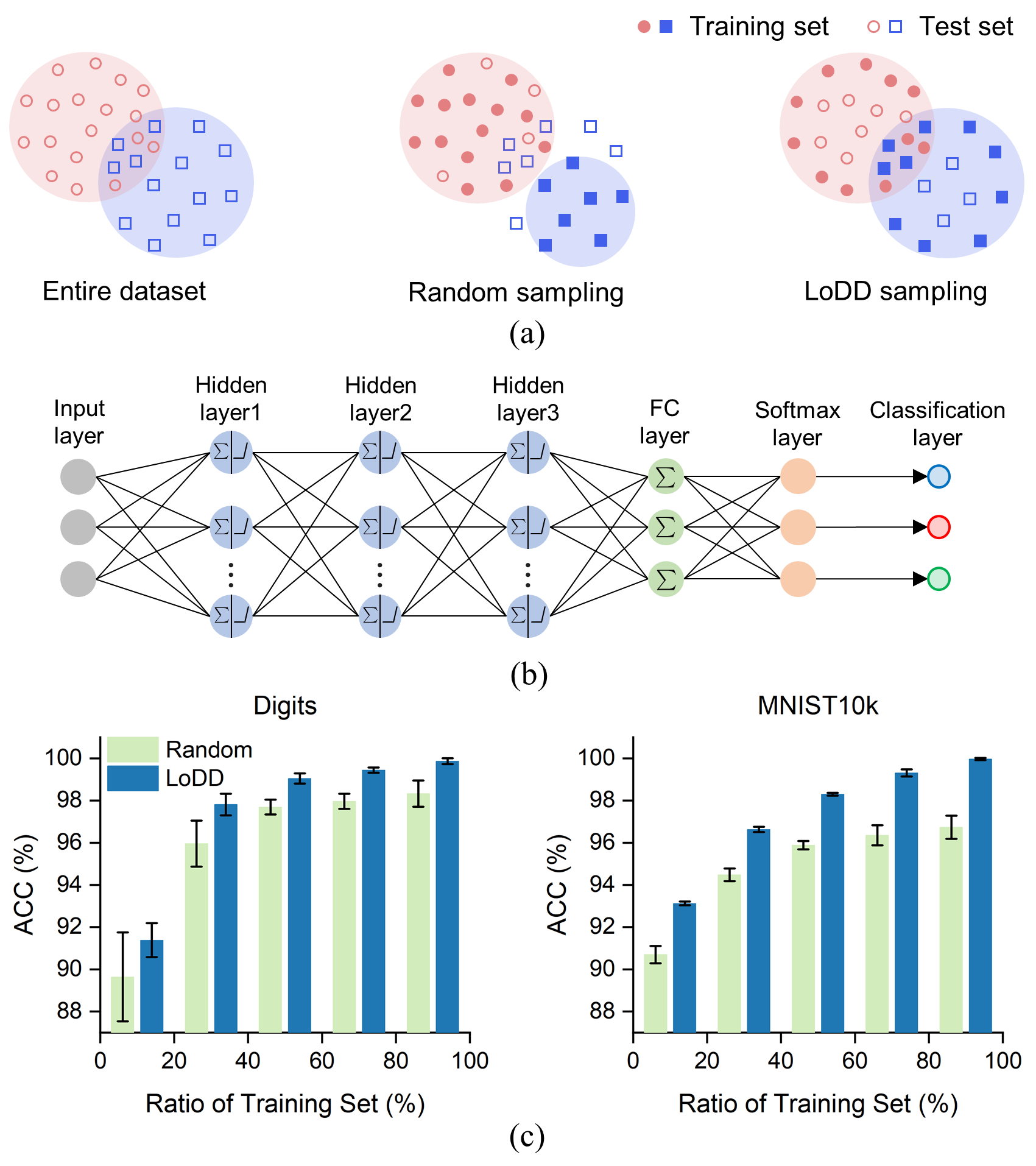}
\caption{Application on training set split. (a) Illustration of the difference between random sampling and LoDD sampling. (b) A DNN model composed of an input layer, three hidden layers, a fully connected layer, a softmax layer, and a classification layer. (b) Accuracy comparison between the DNN model trained on the datasets by random sampling and LoDD-based sampling, where the error bars denote the standard deviations.}
\label{fig10}
\end{figure}

\subsection{Enhancing the Performance of Deep Learning Models} 
The goal of training data split is to select a set of samples to make the deep learning model effectively learn hidden features of the data in training stage. Due to the concise nature, random sampling is the most popular splitting method. However, this approach is unstable and may select non-representative samples, thereby weakening the model’s generalization \coloredcite{r50}. Samples near the cluster boundaries are more likely to be hard samples that affect classification accuracy \coloredcite{r51, r52}, especially when the clusters are very close in the feature space. Therefore, we intend to select the samples with lower LoDD values as the training set, which can represent the original data distribution more comprehensively and bolster the model’s generalization (Fig. \ref{fig10}\textcolor{blue}{(a)})

As shown in Fig. \ref{fig10}\textcolor{blue}{(b)}, a Deep Neural Network (DNN) model was constructed, including an input layer, three hidden layers, a Fully Connected (FC) layer, a softmax layer, and a classification output layer. Following \coloredcite{r53}, we set the number of hidden units to 2/3 of the sum of the sizes of the input and output layers for the first two hidden layers, while the number of hidden units in the third layer was set to half of that value. To introduce nonlinearity, ReLU activation function was used for each hidden layer. The Adam optimizer was adopted, and the maximum number of epochs and the learning rate were set to 100 and 0.001, respectively. We compared the model performances using random sampling and LoDD-based sampling under different ratios of training set from 10\% to 90\% with an interval of 10\% across two handwritten digit datasets, Digits and MNIST10k. In terms of LoDD, parameter $k$ was fixed to 20, and the samples with the smallest $\lbrack ratio*100\%\rbrack$ LoDD values were selected as training samples. To alleviate the impact of randomness, we repeated each experiment for 10 times and reported the average accuracies and standard deviations in Fig. \ref{fig10}\textcolor{blue}{(c)}. The model based on LoDD sampling outperformed that using random sampling in terms of classification accuracy and stability under different ratios of training set. It achieved an average ACC score that is 1.53\% and 3.23\% higher than random sampling under a 90\%-10\% train-test split ratio on Digits and MNIST10k, respectively, and the standard deviation is also 0.5\% lower than that of random sampling. The advantage of LoDD is more pronounced on MNIST10k, since the data distribution in MNIST10k is more complex, and the cluster boundaries are more ambiguous. LoDD can provide boundary cues to make the deep learning models capture the aggregation patterns more accurately, in turn enhancing the performance.
\color{black}

\begin{figure*}[t]
\centering
\includegraphics[width=0.98\linewidth]{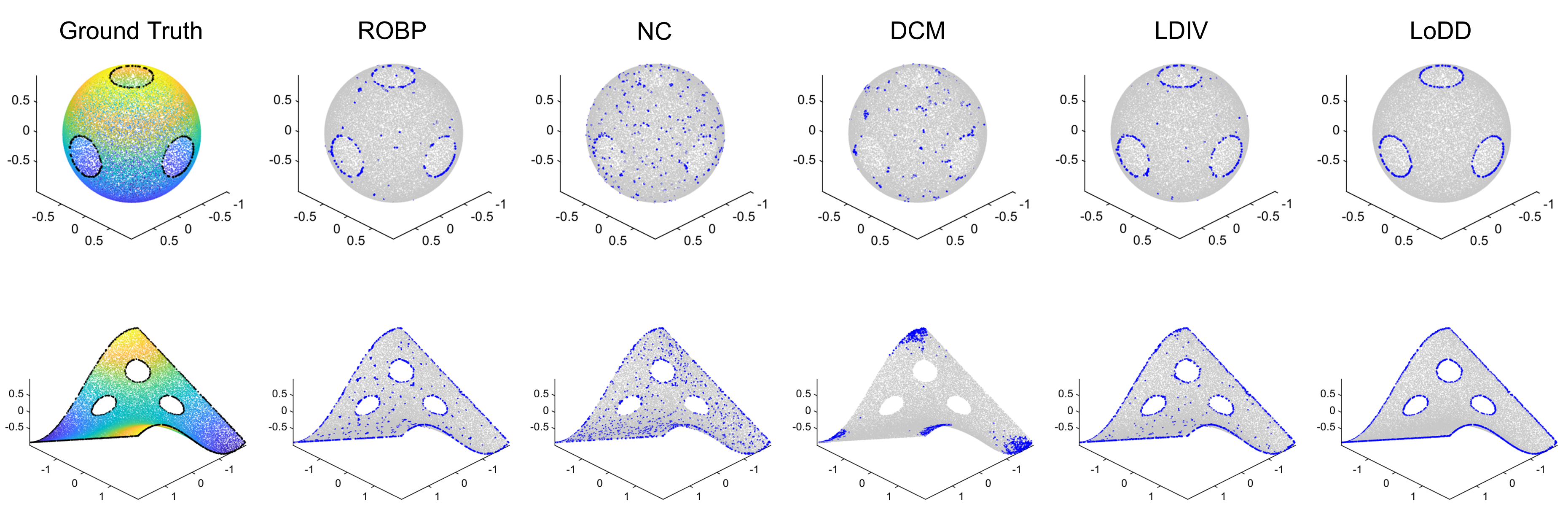}
\caption{Boundaries and holes identified by five boundary point detectors on two 3-D point cloud benchmarks. The black dots in the ground truth and the blue dots in the results denote the true and predicted boundary points, respectively.}
\label{fig11}
\end{figure*}

\subsection{Application on 3-D Point Cloud Data} 
\begin{figure}[t]
\centering
\includegraphics[width=0.9\linewidth]{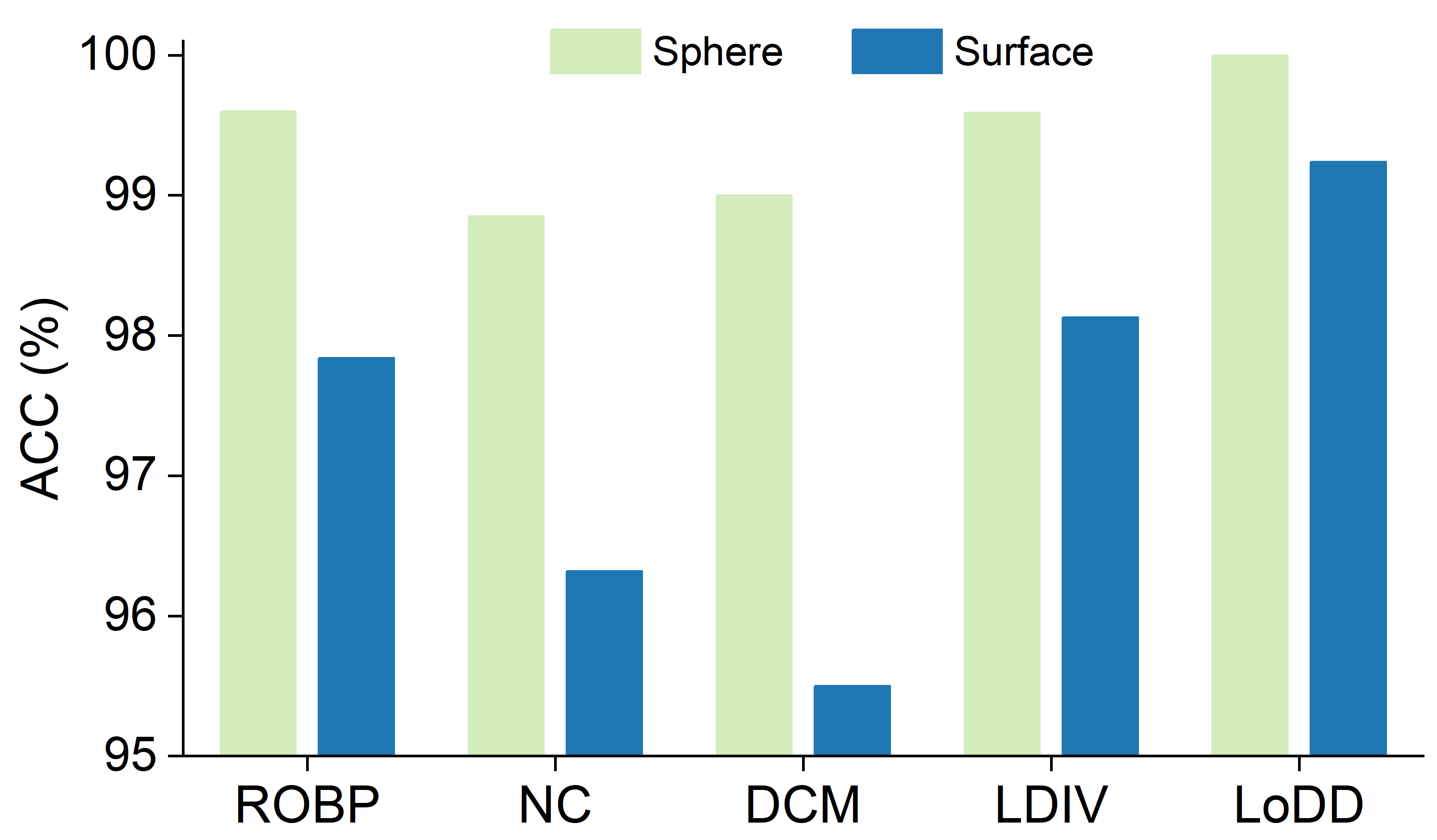}
\caption{A bar plot illustrates the accuracy of five boundary point detectors on two 3-D point cloud benchmarks.}
\label{fig12}
\end{figure}

Boundary and hole detection is the essential step in point cloud processing to serve the hole filling and inpainting of point cloud for better surface reconstruction \coloredcite{r41, r42}. We examined the applicability of LoDD by comparing with four boundary detection baselines on two 3-D point cloud benchmarks. In the first benchmark, 46,299 points form a sphere with three holes. In the second benchmark, 47,046 points generate a surface with three holes and a boundary. We varied $k$ from 10 to 50 and specified $ratio$ as the true proportion of boundary points for all methods. The best results are presented in Fig. \ref{fig11} and the accuracy reported by ACC is shown in Fig. \ref{fig12}. LoDD accurately identified the holes and boundary points and achieved the highest ACC score in both point clouds, even reaching 100\% accuracy in the sphere. The two density-based methods, ROBP and LDIV, performed relatively well, while the drawback of DCM in high-dimensional space was exposed. DCM is defined as the variance of simplex volumes in higher-dimensional spaces, but is insufficient to accurately reflect the uniformity of the KNN distribution when dealing with manifold clusters. Taking the first benchmark as an example, DCM measures the variance of triangular areas in the convex polyhedra formed by the KNN. However, the KNN of each point are almost coplanar since $k \ll n$, resulting in that most of the triangular areas are close to zero and only a few faces have relatively larger areas. If we fix the positions of the vertices of the largest few triangles, changing the distribution of the remaining KNN by assigning them to any vertex positions will hardly affect the variance of the areas. As the result, DCM cannot capture the variation of the KNN distribution in this case.
\section{Conclusion}\label{sec7}
In this paper, we propose a robust boundary point detection method LoDD based on local direction dispersion. It inherits the idea of DCM that considers the distribution uniformity of KNN, but redefines the centrality metric using the projection variances of KNN on PCs. This improvement allows LoDD to more accurately identify boundary points in higher-dimensional spaces, since it considers both the symmetry of the KNN distribution and the balance in various directions. LoDD uses the matrix trace to avoid the computational cost of eigen-decomposition, enabling efficient processing of large-scale high-dimensional datasets. Besides, we develop a method called aLoDD that enables the automatic determination of the parameter $ratio$. The experimental results demonstrated that both LoDD and aLoDD achieved promising outcomes on synthetic datasets and yielded significant accuracy improvements for the K-means algorithm across real-world benchmarks. This method can be used to guide the training set split for deep learning models and detect the boundary and hole points in point cloud data.

Nevertheless, there remains room for further improvement. The globally unified parameter $k$ for KNN may introduce cross-cluster KNN search when handling data with significant density heterogeneity. Particularly for sparse clusters, an overlarge $k$ can lead to query points treating points from other clusters as nearest neighbors, which in turn affects the calculation of local directional dispersion. Future research should focus on developing an adaptive strategy to select an appropriate $k$ based on the local density of each point.
\bibliographystyle{IEEEtran}
\bibliography{ref}

\end{document}